%% file: main.tex
\documentclass[lettersize,journal]{IEEEtran}
\usepackage{amsmath,amsfonts}
\usepackage{algorithmic}
\usepackage{algorithm}
\usepackage{array}
\usepackage{booktabs}
\usepackage{multirow}
\usepackage{enumitem}
\usepackage[caption=false,font=normalsize,labelfont=sf,textfont=sf]{subfig}
\usepackage{textcomp}
\usepackage{stfloats}
\usepackage{url}
\usepackage{verbatim}
\usepackage{graphicx}
\usepackage{cite}
\providecommand{\Description}[1]{}
\def\BibTeX{{\rm B\kern-.05em{\sc i\kern-.025em b}\kern-.08em
    T\kern-.1667em\lower.7ex\hbox{E}\kern-.125emX}}
\usepackage{balance}
\begin{document}
\title{Road-Guided Stochastic Interpolants for Refined Trajectory Generation}

\author{Chenhao Wang, Lisi Chen, Xiaokui Xiao, and Shuo Shang
\thanks{Chenhao Wang, Lisi Chen and Shuo Shang are with School of Computer Science and Engineering, University of Electronic Science and Technology of China, China. (e-mail: chenhao.wang@std.uestc.edu.cn; lchen012@e.ntu.edu.sg; jedi.shang@gmail.com)}
\thanks{Xiaokui Xiao is with School of Computing, National University of Singapore, Singapore. (e-mail: xkxiao@nus.edu.sg)}}


\maketitle

\begin{abstract}
Trajectory generation provides a practical alternative when real-world GPS trajectories are difficult to access due to collection costs and privacy concerns. However, existing GPS trajectory generation models typically map Gaussian noise directly to trajectories, often producing low-quality samples or requiring large denoising models with extensive sampling steps. Moreover, road network information is commonly incorporated through post-processing or conditional embeddings, providing limited fine-grained geometric guidance. Existing methods also lack route-dependent temporal modeling, making it difficult to generate trajectories with both realistic spatial geometry and plausible temporal patterns. To address these limitations, we propose RoadSI, an efficient road-guided stochastic interpolant framework for GPS trajectory generation. RoadSI constructs a learned spatial reference distribution and bridges it to the real spatial distribution with a stochastic interpolant. It improves the reference-real coupling through trajectory re-matching, injects explicit road guidance via a road score, and recovers timestamps through route-conditioned temporal refinement. Experiments on three datasets demonstrate that RoadSI outperforms state-of-the-art baselines in spatial generation metrics and downstream utility, while closely reproducing real temporal distributions. With only 9.79M parameters, RoadSI achieves 1.3--2.9$\times$ faster generation than competing diffusion-based and flow-matching-based methods. Our code is available at \url{https://github.com/chwang0721/RoadSI}.
\end{abstract}


\input{sections/1.introduction}
\input{sections/2.related_work}
\input{sections/3.preliminaries}
\input{sections/4.method}
\input{sections/5.experiments}
\input{sections/6.conclusion}
\balance

\clearpage
\bibliographystyle{IEEEtran}
\bibliography{reference}

\clearpage

\input{sections/appendix}

\end{document}

%% file: sections/1.introduction.tex
\section{Introduction}
GPS trajectories, represented as sequences of timestamped geospatial coordinates, capture rich human mobility patterns and support a variety of urban computing applications, including location prediction~\cite{DBLP:conf/www/FengLZSMGJ18,DBLP:conf/kdd/Xu0ZZ0Z022}, travel time estimation~\cite{DBLP:conf/sigmod/Yuan0BF20,DBLP:journals/pacmmod/LinWHGYLJ23}, and anomalous trajectory detection~\cite{DBLP:conf/icde/Liu0CB20,DBLP:conf/kdd/WangCSJK24}. Despite their broad utility, trajectory-based research remains constrained by the limited availability of large-scale public datasets. Collecting real-world trajectories is costly and labor-intensive~\cite{DBLP:journals/tist/Zheng15}, while privacy concerns and proprietary considerations restrict data sharing~\cite{DBLP:conf/kdd/Cao021,DBLP:journals/tits/ZhuLYY22}. These constraints have motivated the use of synthetic trajectory generation to improve data availability and facilitate reproducible trajectory research.

Trajectory generation aims to produce synthetic trajectories whose distribution approximates that of real trajectories. Early studies primarily rely on rule-based approaches that simulate mobility through handcrafted behavioral assumptions~\cite{DBLP:conf/mobisys/IsaacmanBCMRVW12,jiang2016timegeo,DBLP:journals/tits/YinSFPP18}. Although interpretable, these approaches may struggle to capture the complexity and stochasticity of real-world mobility. Recent studies have increasingly adopted learning-based generative approaches, including GANs~\cite{liu2018trajgans,DBLP:conf/ijcai/OuyangSRY18,DBLP:conf/kdd/FengYXYWL20,DBLP:conf/kdd/Cao021,DBLP:conf/kdd/YuanDWJ022,DBLP:conf/aaai/JiangZWJ23}, VAEs~\cite{DBLP:journals/ijon/ChenXZCFL21,DBLP:conf/kdd/LongWLHWWLLYL23,DBLP:journals/tkde/WangZWJWZY24,DBLP:conf/icdm/LinKS024,DBLP:conf/aaai/CaoZGWD0YSS25}, diffusion models~\cite{DBLP:conf/nips/ZhuYZZY23,DBLP:journals/gis/ChuZWL24,DBLP:journals/tkde/WeiLGLHXBW24,DBLP:conf/kdd/ZhuY0LY00WL24,DBLP:conf/sdm/XuH025,DBLP:conf/www/RaoSJ0025,DBLP:journals/corr/abs-2507-13366}, and flow-matching methods~\cite{DBLP:journals/corr/abs-2603-15009}. By learning trajectory distributions directly from data, these models offer greater flexibility in capturing complex mobility patterns.

\begin{figure}[!t]
    \centering
    \includegraphics[width=0.95\linewidth, trim={0 10mm 0 0}, clip]{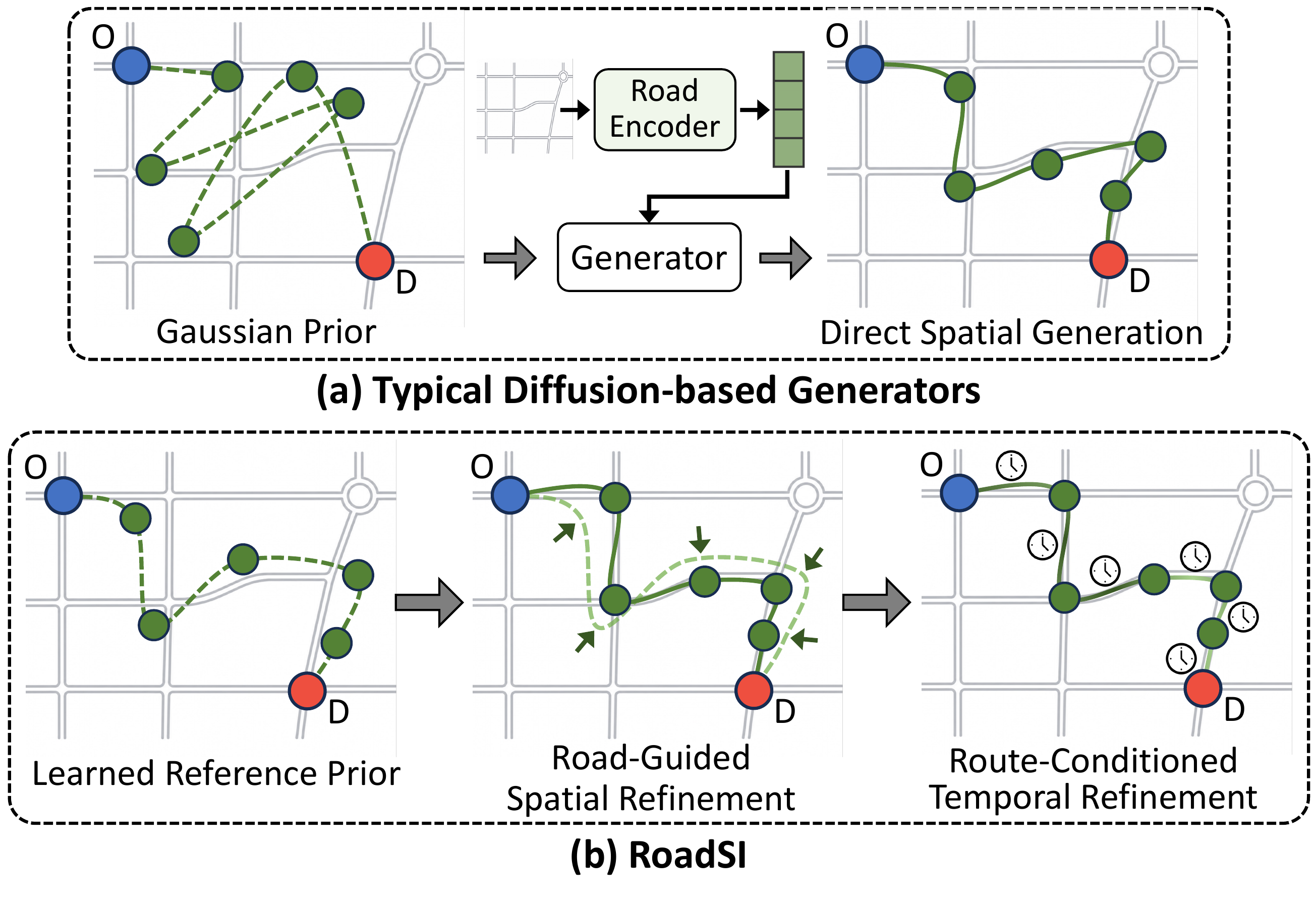}
    \vspace{-7pt}
    \caption{Comparison between typical diffusion-based trajectory generators and RoadSI.}
    \label{fig:intro}
    \vspace{-10pt}
\end{figure}

Despite significant progress, existing learning-based methods still suffer from three deficiencies. First, generation quality and computational efficiency remain difficult to balance. Most existing methods attempt to map an uninformative Gaussian prior directly to real trajectories, creating a large gap to highly structured trajectory distributions. Consequently, lightweight methods (e.g., GANs and VAEs) often produce low-quality samples, while high-performance diffusion models require large denoising backbones and extensive sampling steps. For instance, state-of-the-art methods such as DiffTraj~\cite{DBLP:conf/nips/ZhuYZZY23} and ControlTraj~\cite{DBLP:conf/kdd/ZhuY0LY00WL24} each contain around 30M parameters, and Cardiff~\cite{DBLP:journals/corr/abs-2507-13366} comprises over 50M parameters by incorporating a BART-based~\cite{DBLP:conf/acl/LewisLGGMLSZ20} road segment encoder. Second, road-network guidance is often indirect or inflexible. Road information is commonly incorporated through conditional embeddings~\cite{DBLP:conf/kdd/ZhuY0LY00WL24,DBLP:journals/corr/abs-2507-13366} or post-processing~\cite{DBLP:journals/tkde/WeiLGLHXBW24} (e.g., map-matching), offering limited fine-grained geometric guidance during generation and risking distortion of trajectory geometry. Third, route-dependent temporal dynamics are often insufficiently modeled. Although GPS trajectories contain timestamps, existing methods tend to treat temporal information as auxiliary features or omit it altogether~\cite{DBLP:conf/nips/ZhuYZZY23,DBLP:conf/kdd/ZhuY0LY00WL24,DBLP:journals/corr/abs-2507-13366}, rather than explicitly modeling timestamps conditioned on route geometry. This makes it difficult to generate trajectories with both realistic spatial and temporal patterns.

To address these deficiencies, we propose RoadSI, a road-guided stochastic interpolant framework for GPS trajectory generation. Figure~\ref{fig:intro} illustrates the key differences between RoadSI and typical diffusion-based trajectory generators. The core idea of RoadSI is to (i) construct an informative and well-coupled spatial reference distribution; (ii) refine it into realistic, road-aligned spatial trajectories through a road-guided stochastic interpolant; and (iii) recover route-dependent timestamps through temporal refinement. Implementing RoadSI involves three primary challenges. 
First, the reference distribution must capture mobility patterns while remaining parameter-efficient, and the transport process should ensure a smooth and easily optimizable transition to the real distribution. To address this, we introduce a lightweight diffusion model conditioned on origin--destination pairs for spatial reference generation and a trajectory re-matching strategy to reduce paired trajectory distances and stabilize transport learning. Second, providing fine-grained and explicit road network guidance without post-processing is challenging. We introduce a road-guided refinement mechanism that augments the interpolant's score field with a road score, improving road alignment while preserving physical realism. Third, generating plausible timestamps requires modeling route-dependent temporal dynamics. RoadSI therefore introduces a route-conditioned temporal refinement stage using a log-pace representation to recover spatially consistent timestamps. Evaluations on three datasets against six recent baselines show that RoadSI achieves the best spatial generation performance and the highest efficiency among diffusion- and flow-matching-based methods. Two downstream tasks further demonstrate its superior utility in practical trajectory analytics.

In summary, our main contributions are as follows:
\begin{itemize}[leftmargin=*]
    \item We propose RoadSI, a road-guided stochastic interpolant framework for GPS trajectory generation. To the best of our knowledge, RoadSI is the first work to introduce a learned reference-to-real bridging mechanism for this task.
    \item We construct a learned spatial reference distribution and introduce a trajectory re-matching strategy to improve reference-real couplings, thereby reducing interpolant transport cost and stabilizing the learning process.
    \item We develop a unified refinement process that combines road-guided spatial refinement with route-conditioned temporal refinement, enabling RoadSI to generate trajectories with both road-aligned geometry and plausible temporal patterns.
    \item Experiments on three datasets against six baselines show that RoadSI achieves superior spatial fidelity and downstream utility while generating plausible timestamps. Moreover, with only 9.79M parameters, RoadSI delivers 1.3--2.9$\times$ generation speedups over competing diffusion- and flow-matching-based methods.
\end{itemize}

%% file: sections/2.related_work.tex
\section{Related Work}
\subsection{Trajectory generation.}
Trajectory generation methods can be categorized into two types: rule-based and learning-based~\cite{DBLP:conf/kdd/FengYXYWL20}. Rule-based methods~\cite{DBLP:conf/infocom/SongKJH04,DBLP:conf/mobisys/IsaacmanBCMRVW12,jiang2016timegeo,DBLP:journals/tits/YinSFPP18} rely on predefined behavioral rules such as travel speed, direction, and stop locations. While these methods effectively capture coarse-grained mobility patterns, they struggle to model fine-grained characteristics. Recent studies primarily focus on learning-based methods. GAN-based methods~\cite{liu2018trajgans,DBLP:conf/ijcai/OuyangSRY18,DBLP:conf/kdd/FengYXYWL20,DBLP:conf/kdd/Cao021,DBLP:conf/kdd/YuanDWJ022} typically discretize trajectories into spatial grids, leading to resolution sensitivity. TS-TrajGen~\cite{DBLP:conf/aaai/JiangZWJ23} mitigates this via A*-guided segment-wise generation. VAE-based methods~\cite{DBLP:journals/ijon/ChenXZCFL21,DBLP:conf/kdd/LongWLHWWLLYL23} focus on learning latent mobility representations. Recent advances include TrajSynVAE~\cite{DBLP:journals/tkde/WangZWJWZY24}, which integrates a temporal point process; Geo-CETRA~\cite{DBLP:conf/icdm/LinKS024}, which leverages two encoders with random Fourier features to add spatio-temporal constraints; and HOSER~\cite{DBLP:conf/aaai/CaoZGWD0YSS25}, which incorporates multi-granularity semantics. Recently, diffusion-based methods have emerged as a new paradigm, ranging from GPS-based DiffTraj~\cite{DBLP:conf/nips/ZhuYZZY23} and grid-based TrajGDM~\cite{DBLP:journals/gis/ChuZWL24} to road-segment-based Diff-RNTraj~\cite{DBLP:journals/tkde/WeiLGLHXBW24}. Several works aim to enhance generation quality through specific mechanisms: ControlTraj~\cite{DBLP:conf/kdd/ZhuY0LY00WL24} improves road awareness via conditional generation; AutoSTDiff~\cite{DBLP:conf/sdm/XuH025} adopts an auto-regressive framework; Seed~\cite{DBLP:conf/www/RaoSJ0025} combines a sequence model with a diffusion model for long-range consistency; Traj-Transformer~\cite{DBLP:journals/corr/abs-2510-06291} replaces convolutional denoisers with a Transformer backbone; TrajDLM~\cite{DBLP:journals/corr/abs-2605-10020} performs topology-aware block diffusion over road-segment sequences; Cardiff~\cite{DBLP:journals/corr/abs-2507-13366} utilizes spatial hierarchy modeling; and TrajFlow~\cite{DBLP:journals/corr/abs-2603-15009} applies conditional flow matching. Additionally, LLMs have also been explored for trajectory generation~\cite{DBLP:conf/nips/WangJYWOSK024,DBLP:conf/mdm/LiTLKSZSX25,DBLP:journals/corr/abs-2605-30014}. RoadSI is distinguished by a refinement pipeline that explicitly incorporates road-network guidance into spatial generation and models temporal dynamics conditioned on the refined route, producing road-aligned GPS trajectories with route-consistent timestamps.

\subsection{Diffusion models and stochastic interpolants.}
Diffusion models, first introduced in~\cite{DBLP:conf/icml/Sohl-DicksteinW15}, have become a powerful generative framework achieving strong performance in image~\cite{DBLP:conf/cvpr/RombachBLEO22}, speech~\cite{DBLP:conf/iclr/KongPHZC21}, and text~\cite{DBLP:conf/nips/LiTGLH22} generation. They comprise a forward noising process and a reverse denoising process that reconstructs data from Gaussian noise. DDPM~\cite{DBLP:conf/nips/HoJA20} models the reverse dynamics as a Markov chain, whereas DDIM~\cite{DBLP:conf/iclr/SongME21} accelerates sampling via a non-Markovian formulation. Recent work on stochastic interpolants~\cite{albergo2025stochastic} generalizes diffusion by connecting arbitrary distributions in finite time, enabling non-Gaussian initializations, with BRIDGER~\cite{DBLP:conf/rss/ChenLLCS24} demonstrating their utility in imitation learning. Inspired by these works, RoadSI applies stochastic interpolants in both spatial and temporal spaces: it bridges a learned reference distribution to the real spatial distribution and refines a route-conditioned temporal prior to recover timestamps consistent with the generated route.

%% file: sections/3.preliminaries.tex
\section{Preliminaries}
\subsection{Problem Formulation}

\noindent\textbf{Definition 1 (Trajectory).}
A trajectory is an ordered sequence $\tau=\langle \tau^{(1)}, \tau^{(2)}, \dots, \tau^{(L)} \rangle$ of length $L$, where $\tau^{(i)}=(\ell^{(i)},u^{(i)})$ denotes the $i$-th point in the trajectory. The trajectory can be decomposed into a spatial component and a temporal component: $\tau=(\tau^{\mathrm{sp}},\tau^{\mathrm{tm}})$. The spatial component is given by $\tau^{\mathrm{sp}}=\langle \ell^{(1)},\ell^{(2)},\dots,\ell^{(L)}\rangle$, where $\ell^{(i)}=(\mathrm{lat}^{(i)},\mathrm{lon}^{(i)})$ denotes the $i$-th GPS location in the trajectory. The temporal component is given by $\tau^{\mathrm{tm}}=\langle u^{(1)},u^{(2)},\dots,u^{(L)}\rangle$, where $u^{(i)}$ denotes the timestamp associated with $\ell^{(i)}$.

\noindent\textbf{Definition 2 (Road Network).}
A road network is modeled as a directed graph $G=(V, E)$, where $V$ denotes the set of nodes and $E$ denotes the set of edges. Each node corresponds to a geographic location (e.g., an intersection), and each directed edge denotes a road segment connecting two nodes.

\noindent\textbf{Problem Statement (Trajectory Generation).}
Given real trajectories $\mathcal{T} = \{\tau^1, \tau^2, \dots, \tau^{\left| \mathcal{T} \right|}\}$ and the corresponding road network $G$, the objective is to learn a model that produces synthetic trajectories $\hat{\mathcal{T}} = \{\hat{\tau}^1, \hat{\tau}^2, \dots, \hat{\tau}^{\left| \hat{\mathcal{T}} \right|}\}$ whose distribution matches that of $\mathcal{T}$.

\begin{figure*}[!t]
    \centering
    \Description{Overview of the RoadSI framework.}
    \includegraphics[width=1\linewidth]{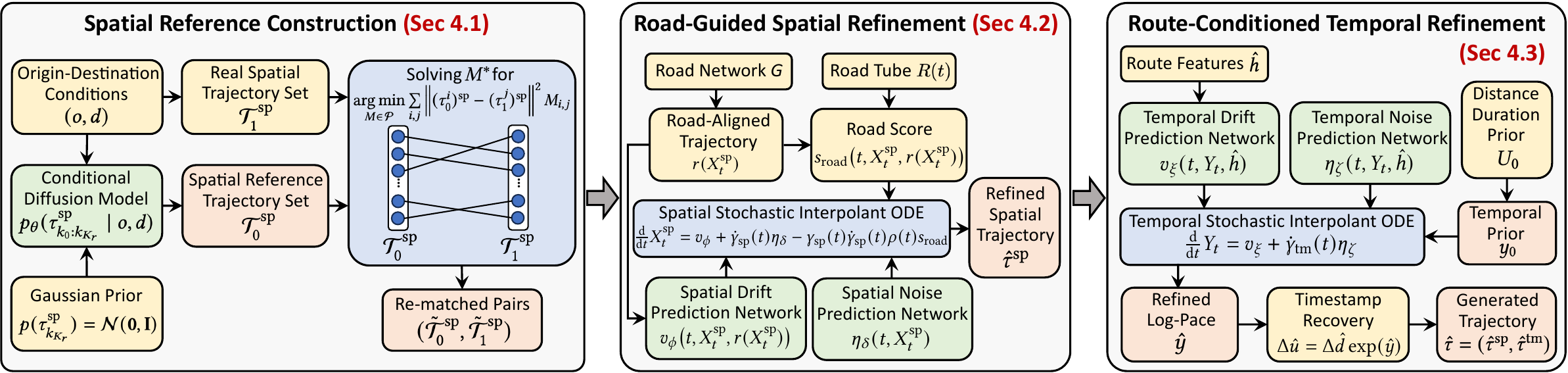}
    \caption{Overview of the RoadSI framework.}
    \label{fig:framework}
\end{figure*}

\subsection{Stochastic Interpolants}
\label{subsec:si}
Unlike conventional diffusion models, which rely on a Gaussian reference distribution, stochastic interpolants~\cite{albergo2025stochastic} transport between arbitrary distributions $\pi_0$ and $\pi_1$. Let $(x_0,x_1)\sim\pi_{0,1}$, where $\pi_{0,1}$ is a coupling of $\pi_0$ and $\pi_1$. A stochastic interpolant is defined by
\begin{equation}
    x_t = I(t, x_0, x_1) + \gamma(t) z, \quad t \in [0, 1],
\end{equation}
where $I$ satisfies $I(0,x_0,x_1)=x_0$ and $I(1,x_0,x_1)=x_1$. Here, $z\sim\mathcal{N}(\mathbf{0},\mathbf{I})$, and the noise schedule $\gamma$ satisfies $\gamma(0)=\gamma(1)=0$ and $\gamma(t)>0$ for all $t\in(0,1)$. For each realization $(x_0,x_1,z)$, the pathwise velocity is given by
\begin{equation}
    \dot{x}_t = \partial_t I(t,x_0,x_1)+\dot{\gamma}(t)z, \quad t\in(0,1).
\end{equation}

The associated state-dependent velocity field is
\begin{equation}
    b(t,x)=\mathbb{E}\!\left[\dot{x}_t\mid x_t=x\right].
\end{equation}

The velocity field $b(t,x)$ defines a probability flow ODE that transports the marginal distribution from $\pi_0$ to $\pi_1$.

%% file: sections/4.method.tex
\section{Methodology}
In this section, we present the RoadSI framework. As shown in Figure~\ref{fig:framework}, the framework consists of three main stages: spatial reference construction, road-guided spatial refinement, and route-conditioned temporal refinement.

\subsection{Spatial Reference Construction}
\subsubsection{Spatial Reference Trajectory Generation}
Spatial reference trajectory generation provides a structured initial distribution $\pi_0^{\mathrm{sp}}$ for the stochastic interpolant, instead of directly using an unstructured Gaussian distribution. To construct this distribution, we train a lightweight conditional diffusion model $p_\theta(\tau_0^{\mathrm{sp}}\mid o,d)$, where $\tau_0^{\mathrm{sp}}$ denotes a generated spatial reference trajectory, and $(o,d)$ is the origin--destination condition extracted from a real trajectory $\tau_1^{\mathrm{sp}}\sim\pi_1^{\mathrm{sp}}$. To distinguish the diffusion steps from the stochastic-interpolant timescale $t\in[0,1]$, we index the reference diffusion process by discrete steps $0 = k_0 < k_1 < \cdots < k_{K_r}$, where $K_r$ is the number of transitions. The corresponding diffusion states are denoted by $\{\tau_{k_i}^{\mathrm{sp}}\}_{i=0}^{K_r}$, where $\tau_{k_0}^{\mathrm{sp}}$ denotes the denoised endpoint and $\tau_{k_{K_r}}^{\mathrm{sp}}\sim\mathcal{N}(\mathbf{0},\mathbf{I})$ is Gaussian noise. During generation, the denoised endpoint is used as the spatial reference trajectory $\tau_0^{\mathrm{sp}}$. The reverse denoising process factorizes as
\begin{equation}
    p_\theta(\tau_{k_0:k_{K_r}}^{\mathrm{sp}} \mid o,d) = p(\tau_{k_{K_r}}^{\mathrm{sp}}) \prod_{i=1}^{K_r} p_\theta(\tau_{k_{i-1}}^{\mathrm{sp}} \mid \tau_{k_i}^{\mathrm{sp}}, o, d),
\end{equation}
where $p(\tau_{k_{K_r}}^{\mathrm{sp}}) = \mathcal{N}(\mathbf{0}, \mathbf{I})$. The model is trained using the noise-prediction objective~\cite{DBLP:conf/nips/HoJA20}:
\begin{equation}
    \mathcal{L}_{\mathrm{ref}}(\theta)
    = \mathbb{E}_{\substack{\tau_1^{\mathrm{sp}}\sim\pi_1^{\mathrm{sp}},\\ i\sim\operatorname{U}\{1,\ldots,K_r\},\\ \epsilon\sim\mathcal{N}(\mathbf{0},\mathbf{I})}}
    \left[ \left\| \epsilon - \epsilon_\theta(\tau_{k_i}^{\mathrm{sp}}, k_i, o, d) \right\|^2 \right].
\end{equation}

During training, $\tau_{k_i}^{\mathrm{sp}}$ is obtained by perturbing $\tau_1^{\mathrm{sp}}$ according to the prescribed noise schedule. After training, for each real trajectory $(\tau_1^j)^{\mathrm{sp}}\in\mathcal{T}_1^{\mathrm{sp}}$, we extract its origin--destination condition $(o^j,d^j)$ and sample $(\tau{_0^j})^{\mathrm{sp}} \sim p_\theta( \tau_0^{\mathrm{sp}} \mid o^j,d^j )$. The generated samples form the spatial reference dataset $ \mathcal{T}_0^{\mathrm{sp}} = \{(\tau_0^j)^{\mathrm{sp}}\}$. Let $p_{\mathrm{OD}}(o,d)$ denote the empirical origin--destination distribution induced by the real trajectory dataset. The overall spatial reference distribution $\pi_0^{\mathrm{sp}}$ is
\begin{equation} \pi_0^{\mathrm{sp}}(\tau) = \int p_\theta(\tau\mid o,d) p_{\mathrm{OD}}(o,d) \,\mathrm{d}(o,d). 
\end{equation} 

Each generated reference trajectory is paired with its corresponding real trajectory through their shared origin--destination condition. The resulting paired datasets $(\mathcal{T}_0^{\mathrm{sp}},\mathcal{T}_1^{\mathrm{sp}})$ are then used to train the stochastic interpolant. Further details of the diffusion process and training objective are provided in Appendix~\ref{ap:reference}.

\begin{figure}[!t]
    \centering
    \Description{Illustration of spatial trajectory re-matching.}
    \includegraphics[width=0.9\linewidth]{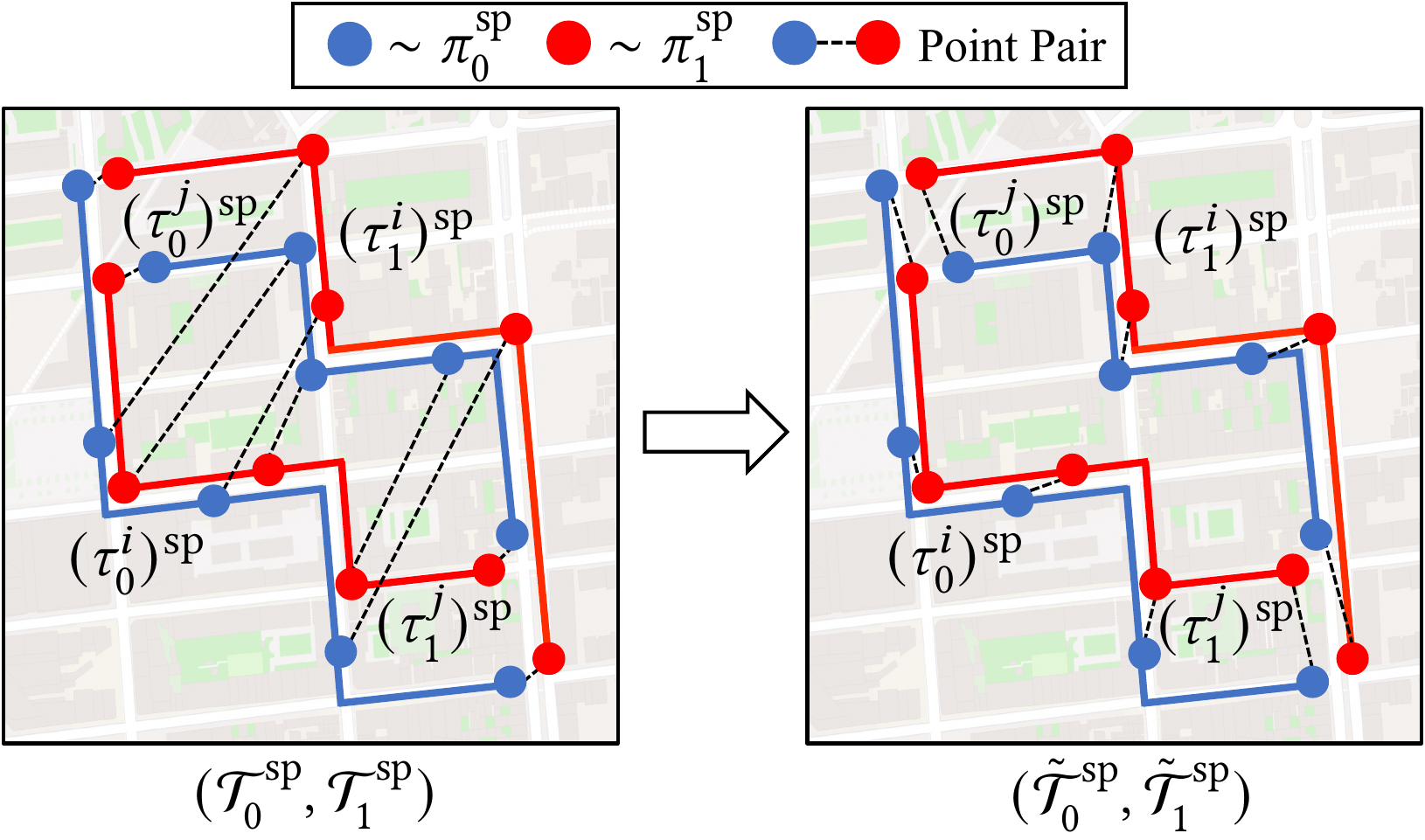}
    \caption{Illustration of spatial trajectory re-matching. The left shows the original trajectory pairs, and the right shows the re-matched trajectory pairs. By re-matching, the transport cost from $\pi_0^{\mathrm{sp}}$ to $\pi_1^{\mathrm{sp}}$ is reduced.}
    \label{fig:re-matching}
\end{figure}

\subsubsection{Spatial Trajectory Re-Matching}
Although spatial reference generation produces origin--destination-conditioned pairs $(\mathcal{T}_0^{\mathrm{sp}},\mathcal{T}_1^{\mathrm{sp}})$, it does not guarantee geometric similarity. As illustrated in Figure~\ref{fig:re-matching}, a reference trajectory $(\tau_0^i)^{\mathrm{sp}}$ may be geometrically dissimilar to its original counterpart $(\tau_1^i)^{\mathrm{sp}}$ while being more closely aligned with another real trajectory $(\tau_1^j)^{\mathrm{sp}}$. Such suboptimal pairings (Figure~\ref{fig:re-matching}, left) require the interpolant to span large geometric discrepancies, potentially driving the interpolation paths through off-road or irrelevant regions. Re-matching (Figure~\ref{fig:re-matching}, right) reduces the pairwise distances between reference and real trajectories, thereby stabilizing training and facilitating the learning of road-aligned geometry.

\noindent\textbf{Interpolant formulation.} To quantify the benefit of re-matching, we specify the stochastic interpolant used in RoadSI. Following Section~\ref{subsec:si}, for a spatial trajectory pair $(\tau_0^{\mathrm{sp}},\tau_1^{\mathrm{sp}})$, we define
\begin{equation}
    \tau_t^{\mathrm{sp}} = (1-t)\tau_0^{\mathrm{sp}} + t\tau_1^{\mathrm{sp}} + \gamma_{\mathrm{sp}}(t) z, \quad t \in [0,1],
\end{equation}
where $z\sim\mathcal{N}(\mathbf{0},\mathbf{I})$, and $\gamma_{\mathrm{sp}}(t) = \lambda_{\gamma}^{\mathrm{sp}} \sqrt{t(1 - t)}$ is a noise schedule satisfying $\gamma_{\mathrm{sp}}(0) = \gamma_{\mathrm{sp}}(1) = 0$. The scaling factor $\lambda_{\gamma}^{\mathrm{sp}}>0$ controls the noise magnitude. The corresponding pathwise velocity is
\begin{equation}
    \dot{\tau}_t^{\mathrm{sp}} = \tau_1^{\mathrm{sp}}-\tau_0^{\mathrm{sp}} +\dot{\gamma}_{\mathrm{sp}}(t)z, \quad t\in(0,1).
\end{equation}

Thus, the velocity consists of a deterministic displacement $(\tau_1^{\mathrm{sp}} - \tau_0^{\mathrm{sp}})$ and a noise-induced term $\dot{\gamma}_{\mathrm{sp}}(t) z$.

\noindent\textbf{Transport cost.} Because $\dot{\gamma}_{\mathrm{sp}}(t)$ is singular at the endpoints, we evaluate the transport cost over the truncated interval $[\varepsilon,1-\varepsilon]$, where $\varepsilon\in(0,\frac{1}{2})$ is a fixed constant:
\begin{equation}
    \mathcal{C} = \int_{\varepsilon}^{1-\varepsilon} \mathbb{E}_z \left[\left\| \tau_1^{\mathrm{sp}}-\tau_0^{\mathrm{sp}} +\dot{\gamma}_{\mathrm{sp}}(t)z \right\|^2 \right] \mathrm{d}t.
\end{equation}

Since $\mathbb{E}_z[z]=\mathbf{0}$, the cross-term vanishes, giving
\begin{equation}
    \mathcal{C} = (1-2\varepsilon)\left\| \tau_1^{\mathrm{sp}}-\tau_0^{\mathrm{sp}}\right\|^2
    + \mathbb{E}_z[\|z\|^2]\int_{\varepsilon}^{1-\varepsilon}\dot{\gamma}_{\mathrm{sp}}^2(t)\,\mathrm{d}t.
\end{equation}

For fixed $\varepsilon$, the second term is finite and independent of the selected pairing. Therefore, minimizing $\mathcal{C}$ over trajectory pairings is equivalent to minimizing $\|\tau_1^{\mathrm{sp}}-\tau_0^{\mathrm{sp}}\|^2$, motivating the re-matching of reference and real trajectories according to their spatial distance.

\noindent\textbf{Minimum-cost re-matching.} To minimize the transport cost, we formulate re-matching as a minimum-weight bipartite matching between the reference set $\mathcal{T}_0^{\mathrm{sp}}$ and the real set $\mathcal{T}_1^{\mathrm{sp}}$, and solve for the assignment matrix $M^*$:
\begin{equation}
    M^* = \arg\min_{M \in \mathcal{P}} \sum_{i,j} \left\| (\tau_0^i)^{\mathrm{sp}} - (\tau_1^j)^{\mathrm{sp}} \right\|^2 M_{i,j},
\end{equation}
where $\mathcal{P}$ is the set of permutation matrices. Here, $M_{i,j}=1$ indicates that the reference spatial trajectory $(\tau_0^i)^{\mathrm{sp}}$ is paired with the real spatial trajectory $(\tau_1^j)^{\mathrm{sp}}$, and $0$ otherwise. This formulation minimizes the total transport cost globally while preserving the marginal distributions $\pi_0^{\mathrm{sp}}$ and $\pi_1^{\mathrm{sp}}$.

We solve the resulting linear assignment problem using the Jonker--Volgenant algorithm~\cite{DBLP:journals/taes/Crouse16}. Since solving the assignment over the entire dataset has $\mathcal{O}(n^3)$ computational complexity, we perform re-matching independently within each batch. This batch-wise strategy provides a scalable approximation to globally optimal matching. We denote the resulting re-matched reference and real trajectory sets by $(\tilde{\mathcal{T}}_0^{\mathrm{sp}}, \tilde{\mathcal{T}}_1^{\mathrm{sp}}) = \left\{\big((\tau_0^i)^{\mathrm{sp}}, (\tau_1^j)^{\mathrm{sp}}\big) \mid M^*_{i,j}=1\right\}$, these better-aligned pairs provide more stable supervision for training the stochastic interpolant.

\subsection{Road-Guided Spatial Refinement}
Given the re-matched trajectory pairs $(\tilde{\mathcal{T}}_0^{\mathrm{sp}}, \tilde{\mathcal{T}}_1^{\mathrm{sp}})$, RoadSI learns the state-dependent velocity field of the spatial stochastic interpolant and augments it with road guidance. For the interpolant defined above, the velocity field can be decomposed as
\begin{equation}
    b(t,x)=v(t,x)+\dot{\gamma}_{\mathrm{sp}}(t)\eta(t,x),
\end{equation}
where
\begin{equation}
    v(t,x)=\mathbb{E}\!\left[\tilde{\tau}_1^{\mathrm{sp}}-\tilde{\tau}_0^{\mathrm{sp}}\mid\tilde{\tau}_t^{\mathrm{sp}}=x\right],
    \quad
    \eta(t,x)=\mathbb{E}\!\left[z\mid\tilde{\tau}_t^{\mathrm{sp}}=x\right].
\end{equation}

The drift field $v(t,x)$ captures the conditional mean displacement between paired trajectories, whereas $\eta(t,x)$ represents the conditional mean of the Gaussian perturbation. Since $\eta(t,x)$ determines the marginal data score, geometric guidance can be naturally introduced at the score level. RoadSI therefore augments the data score with a road score: the former preserves consistency with the learned trajectory distribution, while the latter provides an explicit correction toward the road network.

\noindent\textbf{Data score.} For any $t\in(0,1)$, let $p_t$ denote the marginal density of the spatial interpolant $\tilde{\tau}_t^{\mathrm{sp}}$. The corresponding data score is
\begin{equation} 
    s_{\mathrm{data}}(t,x) = \nabla_x\log p_t(x) = -\gamma_{\mathrm{sp}}^{-1}(t)\eta(t,x).
\end{equation} 

Evaluated at the current interpolant state, it becomes 
\begin{equation}
    s_{\mathrm{data}} \bigl(t,\tilde{\tau}_t^{\mathrm{sp}}\bigr) = -\gamma_{\mathrm{sp}}^{-1}(t) \eta\bigl(t,\tilde{\tau}_t^{\mathrm{sp}}\bigr). 
\end{equation}

The data score points toward regions of higher marginal density, thereby providing distribution-aware guidance during spatial refinement. A detailed derivation is provided in Appendix~\ref{ap:data_score}.

\noindent\textbf{Road score.} To complement the data score with explicit road network geometry, we define a point-wise road score based on a potential. For each spatial point $\tilde{\ell}_t^{(i)}$ in $\tilde{\tau}_t^{\mathrm{sp}}$, we identify its nearest road segment in the road network $G$ and denote the projection onto that segment by $r(\tilde{\ell}_t^{(i)})$. The projection is recomputed for each interpolant state but treated as fixed when differentiating the road potential. To control the permitted deviation from the road network during refinement, we define a time-dependent road-tube radius
\begin{equation}
    R(t) = (1 - t)R_0+ tR_1, \quad t \in [0,1],
\end{equation}
where $R_0$ and $R_{1}$ denote the mean point-to-road distances of the reference and real trajectory datasets, respectively. When $R_1 < R_0$, this schedule progressively tightens the road constraint as $t$ increases. Let
\begin{equation}
    d_t^{(i)} = \big\|\tilde{\ell}_t^{(i)} - r(\tilde{\ell}_t^{(i)})\big\|
\end{equation}
denote the projection distance of the $i$-th point. We define its road potential as
\begin{equation}
    U_{\mathrm{road}}\bigl(t,\tilde{\ell}_t^{(i)}, r(\tilde{\ell}_t^{(i)})\bigr) = \operatorname{softplus}\bigl(d_t^{(i)} - R(t)\bigr),
\end{equation}
where $\operatorname{softplus}(a)=\mathrm{log}(1+e^a)$. The corresponding road score is the negative gradient of this potential:
\begin{equation}
\begin{aligned}
    &s_{\mathrm{road}}\bigl(t, \tilde{\ell}_t^{(i)}, r(\tilde{\ell}_t^{(i)})\bigr) \\
    &\quad = -\nabla_{\tilde{\ell}_t^{(i)}} U_{\mathrm{road}}\bigl(t,\tilde{\ell}_t^{(i)}, r(\tilde{\ell}_t^{(i)})\bigr) \\
    &\quad = - \sigma\bigl(d_t^{(i)} - R(t)\bigr) \frac{\tilde{\ell}_t^{(i)} - r(\tilde{\ell}_t^{(i)})}{d_t^{(i)}}, \quad d_t^{(i)}>0,
\end{aligned}
\end{equation}
where $\sigma(\cdot)$ is the sigmoid function. When $d_t^{(i)}=0$, we set the road score to $\mathbf{0}$. This formulation substantially attenuates the guidance when a point lies well within the road tube and smoothly increases its magnitude as the point moves beyond the tube boundary. The resulting score points toward the road projection, guiding off-road points back toward the road network. The road score for the entire trajectory is denoted by
\begin{equation}
\begin{aligned}
    &s_{\mathrm{road}}\bigl(t,\tilde{\tau}_t^{\mathrm{sp}}, r(\tilde{\tau}_t^{\mathrm{sp}})\bigr) \\
    &\quad = \left\langle
s_{\mathrm{road}}\bigl(t,\tilde{\ell}_t^{(1)},r(\tilde{\ell}_t^{(1)})\bigr),
\dots,
s_{\mathrm{road}}\bigl(t,\tilde{\ell}_t^{(L)},r(\tilde{\ell}_t^{(L)})\bigr)
\right\rangle.
\end{aligned}
\end{equation}

\noindent\textbf{Combined score.} Since the data and road scores capture complementary distributional and geometric information, respectively, we combine them as
\begin{equation}
    s_{\mathrm{comb}}(t,\tilde{\tau}_t^{\mathrm{sp}}) 
    = s_{\mathrm{data}}(t,\tilde{\tau}_t^{\mathrm{sp}})
    + \rho(t)\,s_{\mathrm{road}}\bigl(t,\tilde{\tau}_t^{\mathrm{sp}},r(\tilde{\tau}_t^{\mathrm{sp}})\bigr).
\end{equation}

The time-dependent road-guidance weight is defined as $\rho(t)=\lambda_\rho t\max\{0,2t-1\}$, with $\lambda_{\rho}>0$. This schedule suppresses road guidance during the first half of the transport process and gradually activates it for $t>0.5$. It therefore allows the learned data distribution to determine the global trajectory structure before road guidance enforces local geometric alignment.

\noindent\textbf{Spatial training objective.} We parameterize the drift field $v$ and noise field $\eta$ using two neural networks, $v_\phi\bigl(t,\tilde{\tau}_t^{\mathrm{sp}},r(\tilde{\tau}_t^{\mathrm{sp}})\bigr)$ and
$\eta_\delta(t,\tilde{\tau}_t^{\mathrm{sp}})$, respectively. The drift network is additionally conditioned on the road projection $r(\tilde{\tau}_t^{\mathrm{sp}})$ to incorporate local road geometry.

During training, we sample a re-matched pair $(\tilde{\tau}_0^{\mathrm{sp}},\tilde{\tau}_1^{\mathrm{sp}})$, Gaussian noise $z\sim\mathcal{N}(\mathbf{0},\mathbf{I})$, and an interpolant time $t\sim\operatorname{U}(0,1)$. The drift field is learned by minimizing
\begin{equation}
\begin{aligned}
    \mathcal{L}_v(\phi) &=
    \mathbb{E}\Big[\frac{1}{2}\big\|v_\phi\bigl(t,\tilde{\tau}_t^{\mathrm{sp}},r(\tilde{\tau}_t^{\mathrm{sp}})\bigr) \big\|^2 \\
    &\quad {}-\left\langle \tilde{\tau}_1^{\mathrm{sp}}-\tilde{\tau}_0^{\mathrm{sp}},v_\phi\bigl(t,\tilde{\tau}_t^{\mathrm{sp}},
 r(\tilde{\tau}_t^{\mathrm{sp}})\bigr)\right\rangle\Big].
\end{aligned}
\end{equation}

Similarly, the noise field is learned by minimizing
\begin{equation}
    \mathcal{L}_\eta(\delta)
    =\mathbb{E}\Big[ \frac{1}{2}\left\|\eta_\delta(t,\tilde{\tau}_t^{\mathrm{sp}})\right\|^2
    -\left\langle z,\eta_\delta(t,\tilde{\tau}_t^{\mathrm{sp}})\right\rangle\Big].
\end{equation}

The total spatial training objective is
\begin{equation}
    \mathcal{L}_{\mathrm{sp}}(\phi,\delta) = \mathcal{L}_v(\phi) + \mathcal{L}_\eta(\delta).
\end{equation}

\noindent\textbf{Spatial refinement.}
At inference time, let $X_t^{\mathrm{sp}}$ denote the state of the spatial refinement process. It is initialized by a generated reference trajectory, $X_0^{\mathrm{sp}}\sim\pi_0^{\mathrm{sp}}$. The data score is estimated from the learned noise field and combined with the road score:
\begin{equation}
\begin{aligned}
    s_{\mathrm{comb}}(t, X_t^{\mathrm{sp}}) &= -\gamma_{\mathrm{sp}}^{-1}(t)\eta_\delta(t, X_t^{\mathrm{sp}}) \\
    &\quad {}+ \rho(t) s_{\mathrm{road}}\bigl(t, X_t^{\mathrm{sp}}, r(X_t^{\mathrm{sp}})\bigr).
\end{aligned}
\end{equation}

Starting from the learned probability-flow dynamics, we obtain the road-score-augmented ODE:
\begin{equation}
    \frac{\mathrm{d}}{\mathrm{d}t} X_t^{\mathrm{sp}} = {v}_\phi\big(t, X_t^{\mathrm{sp}}, r(X_t^{\mathrm{sp}})\big) - \gamma_{\mathrm{sp}}(t)\dot{\gamma}_{\mathrm{sp}}(t){s}_{\mathrm{comb}}(t, X_t^{\mathrm{sp}}).
\end{equation}

The combined score balances the learned trajectory distribution while introducing geometric guidance toward the road network. We numerically integrate this ODE using a solver with $K_I$ discrete steps, evaluating the vector field at interior times $t\in(0,1)$. The resulting spatial trajectory is denoted by $\hat{\tau}^{\mathrm{sp}}=X_1^{\mathrm{sp}}$ and subsequently passed to the temporal refinement stage.

\subsection{Route-Conditioned Temporal Refinement}
The road-guided spatial refinement stage determines the trajectory geometry but leaves its temporal dynamics unspecified. RoadSI therefore performs temporal refinement to model the temporal component conditioned on the refined spatial trajectory $\hat{\tau}^{\mathrm{sp}}$.

\noindent\textbf{Temporal representation.}
Directly modeling timestamps is sensitive to arbitrary time origins, whereas raw time intervals do not account for variations in segment length. We therefore represent temporal dynamics using route-dependent log-pace representation. For each segment $i=2,\ldots,L$, we define the segment length and temporal increment as
\begin{equation}
   \Delta d^{(i)} = \|\ell^{(i)}-\ell^{(i-1)}\|, \quad \Delta u^{(i)} = u^{(i)}-u^{(i-1)},
\end{equation}
assuming $\Delta d^{(i)}>0$ and $\Delta u^{(i)}>0$, the segment-level log pace is
\begin{equation}
    y^{(i)} =\log \frac{\Delta u^{(i)}}{\Delta d^{(i)}}, \quad i=2,\dots,L.
\end{equation}

The temporal representation is then given by the segment-level sequence $y=\langle y^{(2)},\dots,y^{(L)}\rangle$. This representation normalizes travel time by segment length, while the logarithmic transformation maps positive pace values to an unconstrained representation space.

\noindent\textbf{Route condition.}
To condition temporal refinement on the spatial route, we construct a route feature sequence $h=\langle h^{(2)},\dots,h^{(L)}\rangle$. The total route distance is $D=\sum_{m=2}^{L}\Delta d^{(m)}$, and the normalized route progress at the end of the $i$-th segment is $c^{(i)}=\sum_{m=2}^{i}\Delta d^{(m)} / D$. For each segment, the route feature vector is
\begin{equation}
    h^{(i)}=\bigl[\Delta d^{(i)}, c^{(i)}, D\bigr].
\end{equation}

The displacement $\Delta d^{(i)}$ captures local movement, while $c^{(i)}$ and $D$ provide relative-position and distance information, respectively.

\noindent\textbf{Temporal prior.}
Similar to spatial refinement, temporal refinement starts from a structured prior rather than Gaussian noise. We construct a route-distance conditioned duration prior by partitioning training trajectories into quantile-based distance bins and estimating the empirical distribution of total durations within each bin. Given a route with total distance $D$, we sample a duration $U_0$ from its corresponding distance bin and distribute it uniformly across the $L-1$ route segments:
\begin{equation}
    \Delta u_0^{(i)} = \frac{U_0}{L-1}, \quad i=2,\ldots,L.
\end{equation}

The resulting temporal prior in the log-pace representation is
\begin{equation}
    y_0^{(i)} =\log\frac{\Delta u_0^{(i)}}{\Delta d^{(i)}}, \quad i=2,\dots,L.
\end{equation}

Thus, $y_0$ provides a duration-realistic temporal prior with uniform segment durations. The temporal branch then refines this prior toward the target representation $y_1$, which is computed from the real timestamps using the log-pace transformation defined above.

\noindent\textbf{Temporal training objective.} Given the condition $h$, temporal prior $y_0$, and target temporal representation $y_1$, RoadSI constructs a stochastic interpolant in the temporal representation space:
\begin{equation}
    y_t=(1-t)y_0+ty_1+{\gamma}_{\mathrm{tm}}(t)z,\quad t\in[0,1],
\end{equation}
where $\gamma_{\mathrm{tm}}(t)=\lambda_{\gamma}^{\mathrm{tm}}\sqrt{t(1-t)}$ is the temporal noise schedule, and $\lambda_{\gamma}^{\mathrm{tm}}>0$ controls its magnitude. We parameterize the temporal drift and noise fields using two route-conditioned neural networks, $v_\xi(t,y_t,h)$ and $\eta_\zeta(t,y_t,h)$. The training objective is:
\begin{equation}
\begin{aligned}
\mathcal{L}_{\mathrm{tm}}(\xi,\zeta)
= \mathbb{E}\Big[
&\frac{1}{2}\left\|v_\xi(t,y_t,h)\right\|^2
-\left\langle y_1-y_0, v_\xi(t,y_t,h) \right\rangle \\
&+\frac{1}{2}\left\|\eta_\zeta(t,y_t,h)\right\|^2
-\left\langle z,\eta_\zeta(t,y_t,h)\right\rangle\Big].
\end{aligned}
\end{equation}

During training, the spatial route is treated as a fixed conditioning input, and gradients are not propagated to the spatial refinement module. The temporal branch therefore learns the conditional temporal dynamics associated with a given spatial route.

\noindent\textbf{Temporal refinement.} At inference time, after obtaining $\hat{\tau}^{\mathrm{sp}}$, we compute its segment lengths $\Delta\hat{d}^{(i)}$, and route condition $\hat{h}$. We then sample a total duration from the corresponding distance bin and construct the temporal prior $y_0$. Let $Y_t$ denote the temporal ODE state, initialized by $Y_0=y_0$. The resulting probability-flow dynamics are given by
\begin{equation} 
\frac{\mathrm{d}}{\mathrm{d}t}Y_t = v_\xi(t,Y_t,\hat{h}) + \dot{\gamma}_{\mathrm{tm}}(t) \eta_\zeta(t,Y_t,\hat{h}).
\end{equation}

We numerically integrate the ODE, evaluating the vector field at interior times $t\in(0,1)$. The generated temporal representation is denoted by $\hat{y}=Y_1$. We decode $\hat{y}$ into segment-level temporal intervals according to
\begin{equation}
    \Delta \hat{u}^{(i)}=\Delta \hat{d}^{(i)}\exp(\hat{y}^{(i)}),\quad i=2,\dots,L.
\end{equation}

The relative timestamps are then recovered by setting $\hat{u}^{(1)}=0$ and $\hat{u}^{(i)}=\sum_{m=2}^{i}\Delta \hat{u}^{(m)}$ for $i=2,\dots,L$. The final generated trajectory is $\hat{\tau}=(\hat{\tau}^{\mathrm{sp}},\hat{\tau}^{\mathrm{tm}})$, where $\hat{\tau}^{\mathrm{tm}}=\langle \hat{u}^{(1)},\hat{u}^{(2)},\dots,\hat{u}^{(L)}\rangle$.

Detailed network architectures for $\epsilon_\theta$, $v_\phi$, $\eta_\delta$, $v_\xi$, and $\eta_\zeta$ are provided in Appendix~\ref{ap:architectures}.

\begin{algorithm}[H]
    \caption{RoadSI training procedure.}
    \label{alg:training}
    \begin{algorithmic}
        \STATE \textbf{Input:} Real trajectory set $\mathcal{T}_1$, road network $G$, batch size $B$, networks $p_\theta$, $v_\phi$, $\eta_\delta$, $v_\xi$ and $\eta_\zeta$, schedules $\gamma_{\mathrm{sp}}$ and $\gamma_{\mathrm{tm}}$
        \STATE \textbf{Output:} Trained parameters $\theta$, $\phi$, $\delta$, $\xi$, $\zeta$
        \STATE Train $p_\theta$ by optimizing $\mathcal{L}_{\mathrm{ref}}(\theta)$ on  $\mathcal{T}_1^{\mathrm{sp}}$;
        \STATE Generate reference spatial set $\mathcal{T}_0^{\mathrm{sp}}$ using $p_\theta$;
        \STATE Construct re-matched pairs $(\tilde{\mathcal{T}}_0^{\mathrm{sp}}, \tilde{\mathcal{T}}_1^{\mathrm{sp}})$ from $(\mathcal{T}_0^{\mathrm{sp}}, \mathcal{T}_1^{\mathrm{sp}})$;
        \FOR{each batch $\{((\tilde{\tau}_0^i)^{\mathrm{sp}},(\tilde{\tau}_1^i)^{\mathrm{sp}})\}_{i=1}^{B}$ from $(\tilde{\mathcal{T}}_0^{\mathrm{sp}}, \tilde{\mathcal{T}}_1^{\mathrm{sp}})$}
            \STATE Sample time step $t_k \sim \operatorname{U}(0,1)$;
            \FOR{each sample $((\tilde{\tau}_0^i)^{\mathrm{sp}},(\tilde{\tau}_1^i)^{\mathrm{sp}})$ in batch}
                \STATE Sample noise $z^i \sim \mathcal{N}(\mathbf{0},\mathbf{I})$;
                \STATE $(\tilde{\tau}_{t_k}^i)^{\mathrm{sp}} \leftarrow (1-t_k)(\tilde{\tau}_0^i)^{\mathrm{sp}} + t_k(\tilde{\tau}_1^i)^{\mathrm{sp}} + \gamma_{\mathrm{sp}}(t_k)z^i$;
                \STATE Compute road-aligned trajectory $r((\tilde{\tau}_{t_k}^i)^{\mathrm{sp}})$ using $G$;
                \STATE $\hat{v}_{\mathrm{sp}}^i \leftarrow v_\phi\big(t_k,(\tilde{\tau}_{t_k}^i)^{\mathrm{sp}},r((\tilde{\tau}_{t_k}^i)^{\mathrm{sp}})\big)$;
                \STATE $\hat{\eta}_{\mathrm{sp}}^i \leftarrow \eta_\delta\big(t_k,(\tilde{\tau}_{t_k}^i)^{\mathrm{sp}}\big)$;
            \ENDFOR
            \STATE Compute $\mathcal{L}_{\mathrm{sp}}$ using $\{(\hat{v}_{\mathrm{sp}}^i,\hat{\eta}_{\mathrm{sp}}^i,z^i)\}_{i=1}^{B}$;
            \STATE Update $\phi,\delta$ via optimization on $\mathcal{L}_{\mathrm{sp}}$;
        \ENDFOR
        \STATE Construct temporal duration prior and temporal triples $\{(y_0^i,y_1^i,h^i)\}$ from $\mathcal{T}_1$;
        \FOR{each batch $\{(y_0^i,y_1^i,h^i)\}_{i=1}^{B}$}
            \STATE Sample time step $t_k \sim \operatorname{U}(0,1)$;
            \FOR{each sample $(y_0^i,y_1^i,h^i)$ in batch}
                \STATE Sample noise $z^i \sim \mathcal{N}(\mathbf{0},\mathbf{I})$;
                \STATE $y_{t_k}^i \leftarrow (1-t_k)y_0^i + t_k y_1^i + \gamma_{\mathrm{tm}}(t_k)z^i$;
                \STATE $\hat{v}_{\mathrm{tm}}^i \leftarrow v_\xi(t_k,y_{t_k}^i,h^i)$;
                \STATE $\hat{\eta}_{\mathrm{tm}}^i \leftarrow \eta_\zeta(t_k,y_{t_k}^i,h^i)$;
            \ENDFOR
            \STATE Compute $\mathcal{L}_{\mathrm{tm}}$ using $\{(\hat{v}_{\mathrm{tm}}^i,\hat{\eta}_{\mathrm{tm}}^i,z^i)\}_{i=1}^{B}$;
            \STATE Update $\xi,\zeta$ via optimization on $\mathcal{L}_{\mathrm{tm}}$;
        \ENDFOR
        \RETURN $\theta$, $\phi$, $\delta$, $\xi$, $\zeta$;
    \end{algorithmic}
\end{algorithm}

\subsection{RoadSI Training and Generation Procedures}
\label{ap:pseudocode}
The pseudocode for the training and generation procedures of RoadSI is detailed in Algorithms~\ref{alg:training} and~\ref{alg:generation}, respectively.

\begin{algorithm}[H]
    \caption{RoadSI generation procedure.}
    \label{alg:generation}
    \begin{algorithmic}
        \STATE \textbf{Input:} Origin-destination condition $(o,d)$, road network $G$, trained networks $p_\theta$, $v_\phi$, $\eta_\delta$, $v_\xi$ and $\eta_\zeta$, final time $t_f \to 1^{-}$, refine steps $K_I$, schedules $\gamma_{\mathrm{sp}}$, $\gamma_{\mathrm{tm}}$ and $\rho$
        \STATE \textbf{Output:} Generated trajectory $\hat{\tau}$
        \STATE Generate reference spatial trajectory $\tau_0^{\mathrm{sp}} \sim p_\theta(\cdot\mid o,d)$;
        \STATE Set $\tilde{\tau}_{t_0}^{\mathrm{sp}} \leftarrow \tau_0^{\mathrm{sp}}$, $t \leftarrow t_0$, and $\Delta_{\mathrm{ode}} \leftarrow (t_f-t_0)/K_I$;
        \WHILE{$t < t_f$}
            \STATE Compute road-aligned trajectory $r(\tilde{\tau}_{t}^{\mathrm{sp}})$ using $G$;
            \STATE $\hat{s}_{\mathrm{comb}} \leftarrow -\gamma_{\mathrm{sp}}^{-1}(t)\eta_\delta(t,\tilde{\tau}_{t}^{\mathrm{sp}}) + \rho(t)s_{\mathrm{road}}\bigl(t,\tilde{\tau}_{t}^{\mathrm{sp}},r(\tilde{\tau}_{t}^{\mathrm{sp}})\bigr)$;
            \STATE $b_t^{\mathrm{sp}} \leftarrow v_\phi\big(t,\tilde{\tau}_{t}^{\mathrm{sp}},r(\tilde{\tau}_{t}^{\mathrm{sp}})\big) - \gamma_{\mathrm{sp}}(t)\dot{\gamma}_{\mathrm{sp}}(t)\hat{s}_{\mathrm{comb}}$;
            \STATE $\tilde{\tau}_{t+\Delta_{\mathrm{ode}}}^{\mathrm{sp}} \leftarrow \tilde{\tau}_{t}^{\mathrm{sp}} + b_t^{\mathrm{sp}}\cdot\Delta_{\mathrm{ode}}$;
            \STATE $t \leftarrow t+\Delta_{\mathrm{ode}}$;
        \ENDWHILE
        \STATE Set $\hat{\tau}^{\mathrm{sp}} \leftarrow \tilde{\tau}_{t_f}^{\mathrm{sp}}$;
        \STATE Construct condition $\hat{h}$ and distances $\{\Delta \hat{d}^{(i)}\}_{i=2}^{L}$ from $\hat{\tau}^{\mathrm{sp}}$;
        \STATE Construct temporal prior ${y}_0$;
        \STATE Set $y_{t_0} \leftarrow {y}_0$ and $t \leftarrow t_0$;
        \WHILE{$t < t_f$}
            \STATE $\hat{s}_{\mathrm{tm}} \leftarrow -\gamma_{\mathrm{tm}}^{-1}(t)\eta_\zeta(t,y_t,\hat{h})$;
            \STATE $b_t^{\mathrm{tm}} \leftarrow v_\xi(t,y_t,\hat{h}) - \gamma_{\mathrm{tm}}(t)\dot{\gamma}_{\mathrm{tm}}(t)\hat{s}_{\mathrm{tm}}$;
            \STATE $y_{t+\Delta_{\mathrm{ode}}} \leftarrow y_t + b_t^{\mathrm{tm}}\cdot\Delta_{\mathrm{ode}}$;
            \STATE $t \leftarrow t+\Delta_{\mathrm{ode}}$;
        \ENDWHILE
        \STATE Set $\hat{y} \leftarrow y_{t_f}$;
        \STATE Set $\hat{u}^{(1)} \leftarrow 0$ and recover $\Delta\hat{u}^{(i)} \leftarrow \Delta \hat{d}^{(i)}\exp(\hat{y}^{(i)})$, $\hat{u}^{(i)} \leftarrow \sum_{m=2}^{i}\Delta\hat{u}^{(m)}$;
        \STATE Construct $\hat{\tau}^{\mathrm{tm}}=\langle\hat{u}^{(1)},\hat{u}^{(2)},\dots,\hat{u}^{(L)}\rangle$;
        \RETURN $\hat{\tau}=(\hat{\tau}^{\mathrm{sp}},\hat{\tau}^{\mathrm{tm}})$;
    \end{algorithmic}
\end{algorithm}

%% file: sections/5.experiments.tex
\begin{table*}[!t]
\centering\small\setlength{\tabcolsep}{0.42mm}
\caption{Spatial Generation Performance of RoadSI and the Baselines on Three Datasets.}
\label{table:performance}
\resizebox{\linewidth}{!}{%
\begin{tabular}{l|ccccc|ccccc|ccccc}
    \toprule
    & \multicolumn{5}{c|}{Shenzhen} & \multicolumn{5}{c|}{Porto} & \multicolumn{5}{c}{Chengdu} \\
    \cmidrule(l){2-16}
    & Dist$\downarrow$ & StepDist$\downarrow$ & Radius$\downarrow$ & Density$\uparrow$ & Pattern$\uparrow$ & Dist$\downarrow$ & StepDist$\downarrow$ & Radius$\downarrow$ & Density$\uparrow$ & Pattern$\uparrow$ & Dist$\downarrow$ & StepDist$\downarrow$ & Radius$\downarrow$ & Density$\uparrow$ & Pattern$\uparrow$ \\
    \midrule
    TrajGAN & 0.02588 & 0.00362 & 0.00270 & 0.63896 & 0.23881 & 0.01109 & 0.00120 & 0.01190 & 0.77047 & 0.41935 & 0.00085 & 0.00041 & 0.00167 & 0.60935 & 0.23881 \\
    TrajVAE & 0.01718 & 0.00284 & 0.00207 & 0.59719 & 0.16606 & 0.02118 & 0.01831 & 0.00502 & 0.76742 & 0.40000 & 0.00072 & 0.00052 & 0.00139 & 0.68218 & 0.29885 \\
    DiffTraj & 0.00127 & 0.00116 & 0.00030 & 0.86568 & 0.51046 & 0.00044 & 0.00054 & 0.00034 & 0.97996 & 0.81690 & 0.00035 & 0.00008 & 0.00015 & \underline{0.93592} & 0.67857 \\
    ControlTraj & \underline{0.00085} & 0.00089 & \underline{0.00017} & 0.87873 & 0.53165 & 0.00039 & 0.00056 & \underline{0.00012} & 0.97874 & \underline{0.84360} & 0.00029 & 0.00012 & \underline{0.00013}  & 0.93074 & 0.71493 \\
    TrajFlow & 0.00127 & 0.00112 & 0.00027 & 0.86362 & 0.52763 & 0.00045 & 0.00051 & 0.00046 & \underline{0.98150} & 0.82734 & 0.00042 & \underline{0.00006} & 0.00017 & 0.93187 & 0.72487 \\
    Cardiff & 0.00109 & \underline{0.00081} & 0.00025 & \underline{0.89651} & \underline{0.55319} & \underline{0.00034} & \underline{0.00036} & 0.00019 & 0.98050 & 0.83019 & \underline{0.00020} & 0.00008 & 0.00024 & 0.93133 & \underline{0.72727} \\
    \midrule
    \textbf{RoadSI} & \textbf{0.00041} & \textbf{0.00063} & \textbf{0.00016} & \textbf{0.92978} & \textbf{0.71493} & \textbf{0.00015} & \textbf{0.00027} & \textbf{0.00007} & \textbf{0.99277} & \textbf{0.89855} & \textbf{0.00016} & \textbf{0.00004} & \textbf{0.00005} & \textbf{0.95599} & \textbf{0.81690} \\
    \textbf{Improve.} & 51.76\% & 22.22\% & 5.88\% & 3.71\% & 29.24\% & 55.88\% & 25.00\% & 41.67\% & 1.15\% & 6.51\% & 20.00\% & 33.33\% & 61.54\% & 2.14\% & 12.32\% \\
    \bottomrule
\end{tabular}%
}
\end{table*}

\begin{table*}[!t]
\centering\small
\setlength{\tabcolsep}{0.27mm}
\caption{Spatial Performance of RoadSI Variants on Three Datasets.}
\label{table:ablation}
\resizebox{\linewidth}{!}{%
\begin{tabular}{l|ccccc|ccccc|ccccc}
    \toprule
    & \multicolumn{5}{c|}{Shenzhen} & \multicolumn{5}{c|}{Porto} & \multicolumn{5}{c}{Chengdu} \\
    \cmidrule(l){2-16}
    & Dist$\downarrow$ & StepDist$\downarrow$ & Radius$\downarrow$ & Density$\uparrow$ & Pattern$\uparrow$ & Dist$\downarrow$ & StepDist$\downarrow$ & Radius$\downarrow$ & Density$\uparrow$ & Pattern$\uparrow$ & Dist$\downarrow$ & StepDist$\downarrow$ & Radius$\downarrow$ & Density$\uparrow$ & Pattern$\uparrow$ \\
    \midrule
    RoadSI-Gaussian & 0.29972 & 0.03694 & 0.22601 & 0.56044 & 0.16606 & 0.59204 & 0.15439 & 0.08899 & 0.86988 & 0.54237 & 0.56855 & 0.06299 & 0.04657 & 0.77290 & 0.40964 \\
    RoadSI-Shortest & 0.00090 & 0.00255 & 0.00060 & \underline{0.92482} & 0.62009 & 0.00491 & 0.00706 & 0.00199 & 0.96376 & 0.73973 & \underline{0.00020} & 0.00014 & 0.00015 & 0.93555 & 0.66667 \\
    RoadSI-CVAE & 0.00191 & 0.00261 & 0.00035 & 0.86389 & 0.65487 & 0.01198 & 0.01068 & 0.00088 & 0.95893 & 0.72727 & 0.00374 & 0.00091 & 0.00043 & \underline{0.93937} & 0.69058 \\
    Reference-Only & 0.00418 & 0.00106 & 0.00044 & 0.82402 & 0.43902 & 0.00090 & 0.00114 & 0.00046 & 0.95227 & 0.77778 & 0.00038 & 0.00034 & 0.00048 & 0.87201 & 0.58621 \\
    RoadSI w/o TR & \textbf{0.00041} & 0.00086 & \underline{0.00023} & 0.90938 & 0.64317 & \underline{0.00020} & 0.00041 & \underline{0.00010} & 0.98511 & 0.85714 & 0.00021 & \underline{0.00006} & 0.00013 & 0.93411 & 0.70270 \\
    RoadSI w/o RS & \underline{0.00059} & \underline{0.00070} & 0.00025 & 0.91785 & \underline{0.67857} & 0.00028 & \underline{0.00036} & 0.00013 & \underline{0.98868} & \underline{0.87081} & 0.00029 & 0.00007 & \underline{0.00012} & 0.92818 & \underline{0.71493} \\
    \midrule
    \textbf{RoadSI} & \textbf{0.00041} & \textbf{0.00063} & \textbf{0.00016} & \textbf{0.92978} & \textbf{0.71493} & \textbf{0.00015} & \textbf{0.00027} & \textbf{0.00007} & \textbf{0.99277} & \textbf{0.89855} & \textbf{0.00016} & \textbf{0.00004} & \textbf{0.00005} & \textbf{0.95599} & \textbf{0.81690} \\
    \bottomrule
\end{tabular}%
}
\end{table*}

\section{Experiments}
\subsection{Datasets}
\label{ap:datasets}
\textbf{Shenzhen}\footnote{https://people.cs.rutgers.edu/dz220/data.html}~\cite{DBLP:conf/iccps/ZhangZZH15} contains GPS trajectories collected from 11,100 taxis in Shenzhen, China, on October 22, 2013, with an average sampling interval of 15 seconds. \textbf{Porto}\footnote{https://www.kaggle.com/c/pkdd-15-predict-taxi-service-trajectory-i/data} consists of trajectories from 442 taxis in Porto, Portugal, recorded between January 2013 and June 2014, with a fixed sampling interval of 15 seconds. \textbf{Chengdu}\footnote{https://js.dclab.run/v2/cmptDetail.html?id=175} includes trajectories from approximately 14,000 taxis in Chengdu, China, collected from August 3 to August 30, 2014, with an average sampling interval of 30 seconds. In our experiments, we use a one-day subset of the Chengdu dataset.

For all three datasets, we select trajectories within a specified latitude–longitude region. We then filter samples based on sequence length, retaining trajectories with lengths in $[20, 200]$. Table~\ref{table:datasets} summarizes the statistics of the preprocessed datasets. We then resample all the trajectories to a fixed length of 200 using linear interpolation. We obtain the road networks from OpenStreetMap\footnote{https://www.openstreetmap.org}.

\begin{table}[!t]\centering\small
\caption{Dataset Statistics.}
\label{table:datasets}\setlength{\tabcolsep}{1mm}
\resizebox{\linewidth}{!}{%
\begin{tabular}{lcccc}
    \toprule
    Dataset & Trajectories & Points & Latitude Range & Longitude Range\\
    \midrule
    Shenzhen  & 56,456 & 2,744,360 & [22.48, 22.58] & [113.9, 114.1]\\
    Porto & 1,023,159 & 46,798,673 & [41.14, 41.186] & [-8.69, -8.55]\\
    Chengdu & 340,868 & 19,016,685 & [30.56, 30.79] & [103.92, 104.21]\\
    \bottomrule
\end{tabular}%
}
\end{table}

\subsection{Baselines}
\label{ap:baselines}
\subsubsection{Spatial Generation Baselines}
\begin{itemize}[leftmargin=*]
    \item \textbf{TrajGAN}~\cite{liu2018trajgans}: TrajGAN is a GAN-based framework designed to generate synthetic trajectories for geo-privacy protection. It leverages an encoder-decoder architecture to transform trajectories into dense representations. A GAN generator learns to produce synthetic trajectory representations from random noise, while a discriminator distinguishes real from generated samples.
    \item \textbf{TrajVAE}~\cite{DBLP:journals/ijon/ChenXZCFL21}: TrajVAE is an LSTM-based VAE that encodes trajectories into a latent Gaussian space, and decodes latent samples to reconstruct GPS sequences.
    \item \textbf{DiffTraj}\footnote{https://github.com/YasoZ/DiffTraj}~\cite{DBLP:conf/nips/ZhuYZZY23}: DiffTraj is a diffusion-based trajectory generator. It employs a trajectory-specific UNet as the denoising network and conditions the reverse diffusion process on spatial–temporal factors such as trip origin, destination, and motion attributes. By modeling GPS sequences directly in the diffusion framework, DiffTraj achieves high-fidelity synthetic trajectories that preserve real-world spatial–temporal distributions.
    \item \textbf{ControlTraj}\footnote{https://github.com/Yasoz/ControlTraj}~\cite{DBLP:conf/kdd/ZhuY0LY00WL24}: ControlTraj is an extension of DiffTraj that incorporates road-topology information into the diffusion process via learned road-segment embeddings. It integrates a masked road-segment autoencoder and a geography-aware UNet, enabling controllable and high-fidelity trajectory generation.
    \item \textbf{TrajFlow}\footnote{https://github.com/ZeroCSIS/TrajFlow}~\cite{DBLP:journals/corr/abs-2603-15009}: TrajFlow is a conditional flow-matching framework for GPS trajectory generation. It applies RDP-based trajectory harmonization to obtain a compact trajectory representation and learns a continuous velocity field that transports Gaussian noise to the normalized trajectory distribution. Generated samples are obtained by integrating the learned ODE and are subsequently reconstructed in geographic coordinates.
    \item \textbf{Cardiff}\footnote{https://github.com/urban-mobility-generation/Cardiff}~\cite{DBLP:journals/corr/abs-2507-13366}: Cardiff is a coarse-to-fine cascaded diffusion framework for trajectory generation. It exploits the hierarchical structure of urban mobility by first generating discrete road-segment trajectories in a latent space using a diffusion Transformer, and then using them as conditions for a GPS-level denoising network.
\end{itemize}

\subsubsection{Temporal Generation Baselines}
\begin{itemize}[leftmargin=*]
    \item \textbf{Constant Speed}: This non-parametric baseline assigns timestamps to each generated spatial trajectory using a fixed speed estimated from the training set. The resulting time intervals are proportional to the corresponding segment distances, yielding a constant speed along the trajectory.
    \item \textbf{Conditional GRU}~\cite{DBLP:conf/emnlp/ChoMGBBSB14}: This baseline uses a GRU to autoregressively model the temporal sequence conditioned on the generated spatial route.
    \item \textbf{Conditional DDPM}~\cite{DBLP:conf/nips/HoJA20}: This baseline models the temporal sequence with a denoising diffusion probabilistic model. Its denoising network is conditioned on the generated spatial route and produces the temporal sequence through iterative reverse diffusion from Gaussian noise.
    \item \textbf{Duration Prior}: This non-parametric baseline partitions the training trajectories into route-distance bins and estimates an empirical distribution of total durations for each bin. Given a generated route, it samples a total duration from the corresponding bin and distributes it uniformly over the trajectory intervals.
\end{itemize}

\subsection{Evaluation Metrics}
\label{ap:metrics}
We use Jensen-Shannon divergence (JSD) to quantify the distributional discrepancy between generated and real trajectories. The JSD between two distributions $P$ and $Q$ is defined as:
\begin{equation}
    \mathrm{JSD} = \frac{1}{2} \mathrm{KL}(P \parallel M) + \frac{1}{2} \mathrm{KL}(Q \parallel M),
\end{equation}
where $\mathrm{KL}(\cdot\parallel\cdot)$ denotes the KL divergence, and $M=\frac{1}{2}(P + Q)$ is the mixture distribution of $P$ and $Q$.

The metrics used in our experiments are described as follows:
\begin{itemize}[leftmargin=*]
    \item \textbf{Distance Error (Dist)}: This metric measures the discrepancy in travel distance between real and generated trajectories. For each trajectory, the travel distance is computed as the sum of Euclidean distances between successive points. We then calculate the JSD between the travel-distance distributions of generated and real trajectories.
    \item \textbf{Step Distance Error (StepDist)}: This metric measures the discrepancy in step-wise distances between real and generated trajectories. The step-wise distance is defined as the Euclidean distance between two consecutive points, and we compute the JSD between the step-wise distance distributions of the two datasets.
    \item \textbf{Radius Error (Radius)}: This metric measures the spatial range of a trajectory by computing the root mean square distance of all points from the trajectory’s centroid. We compare the distributions of radius values between real and generated trajectories using JSD.
    \item \textbf{Density Similarity (Density)}: This metric evaluates the similarity of spatial point density between real and generated trajectories. We partition each city into a $\textrm{100m}\times\textrm{100m}$ grid and count the number of trajectory points falling in each cell. Cosine similarity is then computed between the grid-density distributions of the real and generated trajectories.
    \item \textbf{Pattern Score (Pattern)}: This metric evaluates whether the generated trajectories reproduce the most frequently visited spatial regions. We partition each city into a $\textrm{100m}\times\textrm{100m}$ grid and identify the top 100 most visited grid cells in the real dataset as ground truth and compute the F1 score between the top 100 grid cells from the real and generated datasets.
    \item \textbf{Travel Time Error (TravelTime)}: This metric measures the discrepancy in trip duration between real and generated trajectories. For each trajectory, the travel time is computed as the time difference between the final and initial timestamps. We then calculate the JSD between the travel-time distributions of generated and real trajectories.
    \item \textbf{Time Interval Error (TimeInterval)}: This metric measures the discrepancy in the time intervals between consecutive trajectory points. For each trajectory, we compute $\Delta u^{(i)}=u^{(i)}-u^{(i-1)}$ for $i=2,\dots,L$. We then calculate the JSD between the time-interval distributions of generated and real trajectories.
    \item \textbf{Speed Error (Speed)}: This metric measures the discrepancy in movement speed between real and generated trajectories. For each trajectory segment, the speed is computed as the distance between two consecutive points divided by the corresponding time interval. We then calculate the JSD between the speed distributions of generated and real trajectories.
\end{itemize}

\subsection{Implementation Details}
\label{ap:impl}
RoadSI is implemented using Python 3.10 and PyTorch 2.6.0. All experiments are conducted on a server equipped with a 40-core Intel Xeon CPU, four NVIDIA RTX 4090 GPUs, and 256 GB of RAM. The base channel size is set to 128 for the reference generation network $p_\theta$ and the spatial and temporal drift/noise prediction networks. The spatial networks $v_\phi$ and $\eta_\delta$ each contain five residual blocks, while the temporal networks $v_\xi$ and $\eta_\zeta$ each contain two.

We use AdamW with a global batch size of 1024. For Shenzhen, the learning rate is set to 0.0006 and the models are trained for 2000 epochs. For Porto and Chengdu, the learning rate is set to 0.0003, with 500 and 1000 training epochs, respectively. The reference diffusion model uses 500 training steps with a linear noise schedule in which $\beta_k$ ranges from $0.0001$ to $0.04$, followed by 50 DDIM sampling steps. Both the spatial and temporal stochastic interpolants are integrated using 50 steps.

For trajectory re-matching, we apply the Jonker--Volgenant algorithm within batches of size $N_s$, $\lfloor N_p/16 \rfloor$, and $\lfloor N_c/8 \rfloor$ for Shenzhen, Porto, and Chengdu, respectively. The temporal duration prior is estimated using eight quantile-based route-distance bins. We set $\lambda_{\gamma}^{\mathrm{sp}}=0.3$, $\lambda_{\gamma}^{\mathrm{tm}}=0.05$, and $\lambda_{\rho}=0.5$ as the default setting. All experiments are conducted with three different random seeds, and we report the mean performance.

\subsection{Overall Generation Performance}
Table~\ref{table:performance} compares the spatial generation performance of RoadSI with the baselines. For fairness, we fix the total number of sampling steps at 100 for each diffusion-based and flow-matching-based method. For RoadSI, 50 steps are allocated to the reference trajectory generator and the remaining 50 steps to the stochastic interpolant.

RoadSI achieves the best performance across all datasets. On the macro-level metrics (Density and Pattern), its average relative improvements over the second-best results are 16.47\%, 3.83\%, and 7.23\% on Shenzhen, Porto, and Chengdu, respectively. On the micro-level metrics (Dist, StepDist, and Radius), the corresponding average improvements are 26.62\%, 40.85\%, and 38.29\%. These gains demonstrate that RoadSI improves both fidelity to global spatial distributions and fine-grained geometric realism. We attribute them to trajectory re-matching, which constructs better aligned reference-real pairs, and road-guided trajectory refinement, which provides explicit geometric guidance throughout generation.

Among the baselines, diffusion-based and flow-matching-based methods outperform GAN-based and VAE-based methods. ControlTraj or Cardiff ranks second in most comparisons, indicating the benefit of incorporating road-network information. Nevertheless, RoadSI maintains a clear advantage, particularly on Dist, StepDist, and Radius. This advantage suggests that explicitly incorporating road geometry into the refinement provides more direct and effective guidance than relying solely on road-conditioned generation.

Table~\ref{table:temporal} reports the temporal generation performance of RoadSI and the four temporal baselines. RoadSI achieves the best results in eight of the nine dataset-metric combinations and ranks second in the remaining one. In particular, RoadSI consistently obtains the lowest TravelTime and Speed errors across all three datasets. Compared with the corresponding second-best results, it reduces TravelTime error by 6.80\%, 20.93\%, and 46.81\%, and Speed error by 33.33\%, 55.78\%, and 76.92\% on Shenzhen, Porto, and Chengdu, respectively. For TimeInterval, RoadSI reduces the error by 40.00\% on Shenzhen and 63.94\% on Chengdu, whereas Duration-Prior achieves the best result on Porto. Overall, these results indicate that route-conditioned temporal refinement effectively models both trip-level travel duration and fine-grained local movement dynamics.

\begin{table}[!t]
\centering\small\setlength{\tabcolsep}{0.9mm}
\caption{Temporal Generation Performance Comparison}
\resizebox{\linewidth}{!}{%
\begin{tabular}{llccc}
    \toprule
    Dataset & Method & TravelTime $\downarrow$ & TimeInterval $\downarrow$ & Speed $\downarrow$ \\
    \midrule
    \multirow[c]{5}{*}{Shenzhen}
     & Constant Speed & 0.01303 & 0.00118 & 0.78546 \\
     & Cond GRU & 0.00459 & 0.00055 & 0.00011 \\
     & Cond DDPM & 0.00273 & 0.00037 & 0.00009 \\
     & Duration Prior & \underline{0.00103} & \underline{0.00025} & \underline{0.00003} \\
     & RoadSI & \textbf{0.00096} & \textbf{0.00015} & \textbf{0.00002} \\
    \midrule
    \multirow[c]{5}{*}{Porto}
     & Constant Speed & 0.04490 & 0.03400 & 0.74078 \\
     & Cond GRU & 0.01416 & 0.01907 & 0.00758 \\
     & Cond DDPM & 0.00353 & 0.01268 & 0.00451 \\
     & Duration Prior & \underline{0.00043} & \textbf{0.00066} & \underline{0.00441} \\
     & RoadSI & \textbf{0.00034} & \underline{0.00106} & \textbf{0.00195} \\
    \midrule
    \multirow[c]{5}{*}{Chengdu}
     & Constant Speed & 0.00865 & 0.00717 & 0.80814 \\
     & Cond GRU & 0.00542 & 0.00604 & 0.00083 \\
     & Cond DDPM & 0.00243 & 0.00352 & 0.00057 \\
     & Duration Prior & \underline{0.00047} & \underline{0.00330} & \underline{0.00013} \\
     & RoadSI & \textbf{0.00025} & \textbf{0.00119} & \textbf{0.00003} \\
    \bottomrule
\end{tabular}%
}
\label{table:temporal}
\end{table}

\begin{figure*}[!t]
\centering
\Description{Model efficiency comparison.}
\subfloat[Shenzhen.]{%
  \includegraphics[width=0.32\textwidth]{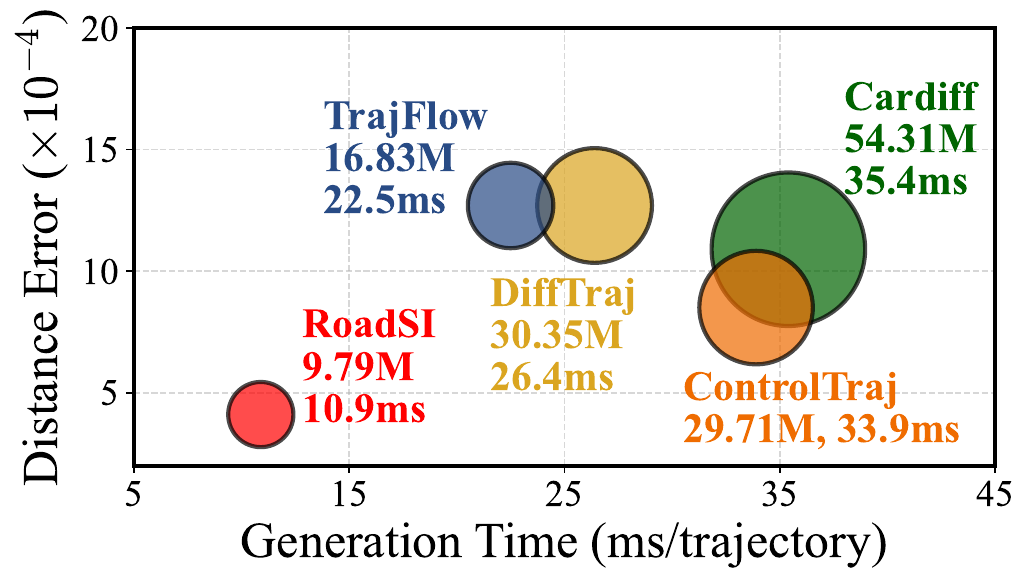}}
\quad
\subfloat[Porto.]{%
  \includegraphics[width=0.32\textwidth]{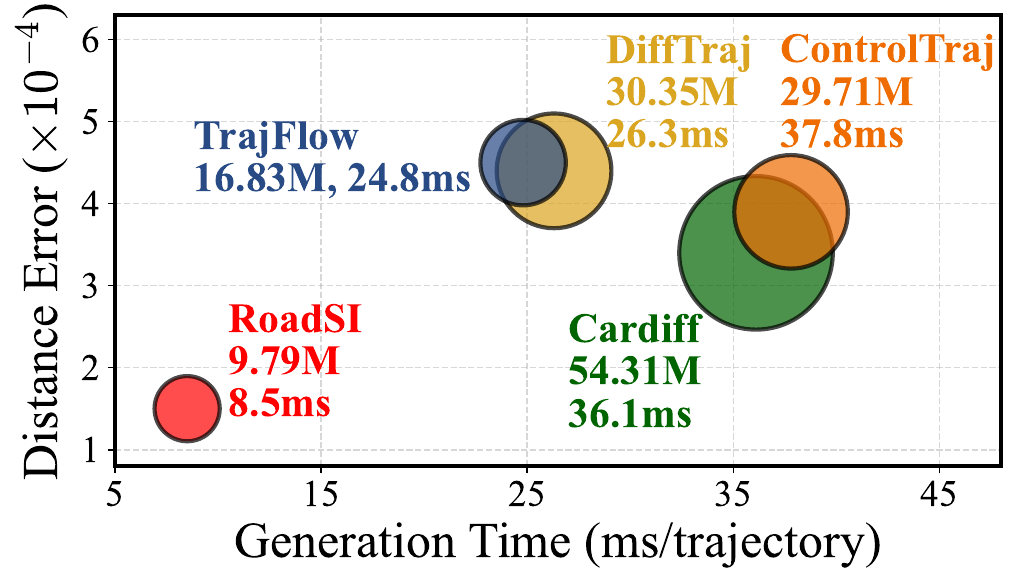}}
\quad
\subfloat[Chengdu.]{%
  \includegraphics[width=0.32\textwidth]{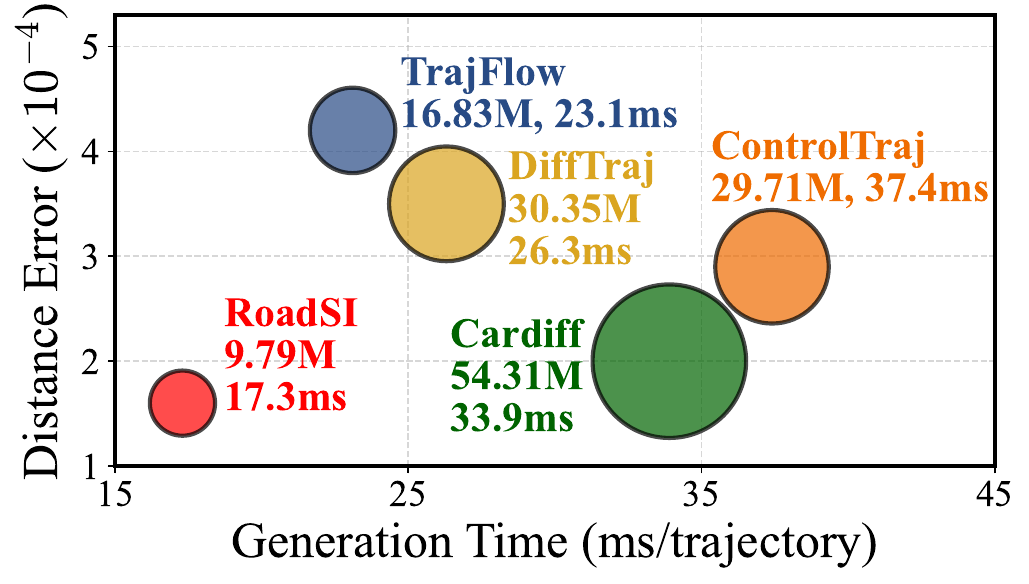}}
\caption{Comparison of generation quality and runtime, bubble area indicates model size.}
\label{fig:efficiency}
\end{figure*}

\subsection{Ablation Study}
Table~\ref{table:ablation} evaluates six variants of RoadSI. (1) RoadSI-Gaussian replaces the learned reference distribution with a standard Gaussian distribution. (2) RoadSI-Shortest uses the shortest path between origin--destination pair as the reference trajectory. (3) RoadSI-CVAE employs a conditional VAE to generate reference trajectories. (4) Reference-Only directly uses the output of the reference generator without subsequent refinement. (5) RoadSI w/o TR removes trajectory re-matching and retains the original reference-real trajectory pairs. (6) RoadSI w/o RS removes the road score, leaving the refinement process guided only by the learned stochastic interpolant.

\noindent\textbf{Impact of the reference distribution.} RoadSI-Gaussian performs substantially worse than the other variants, highlighting the importance of an informative reference distribution. RoadSI-Shortest and RoadSI-CVAE both markedly outperform RoadSI-Gaussian, but they remain inferior to the full RoadSI. These results suggest that shortest-path references do not adequately capture the diversity of real mobility patterns, while CVAE-generated references provide a less effective initialization for subsequent refinement. In comparison, the learned reference generator in RoadSI produces more realistic and informative initial trajectories, thereby reducing the difficulty of transforming reference samples into real trajectories.

\noindent\textbf{Effectiveness of the interpolant components.} Reference-Only underperforms both the complete RoadSI and diffusion-based and flow-matching-based baselines, indicating that the learned reference distribution alone is insufficient and the subsequent stochastic-interpolant refinement is necessary. Removing trajectory re-matching degrades performance in 14 of the 15 dataset-metric combinations. This result confirms the importance of constructing well aligned reference-real pairs, as re-matching reduces geometric mismatch. Removing the road score degrades performance in all dataset-metric combinations, indicating that explicit road-guided refinement consistently benefits spatial trajectory generation.

\subsection{Efficiency Study}
We compare RoadSI with diffusion-based and flow-matching-based baselines in spatial generation quality, runtime, and parameter count. Runtime is measured for spatial generation only for RoadSI. We fix the total sampling steps at 100 and the batch size at 1024, using a single RTX 4090 GPU.

As shown in Figure~\ref{fig:efficiency}, RoadSI achieves both the lowest distance error and the shortest generation time across all three datasets. RoadSI contains 9.79M parameters (6.23M reference, 2.67M spatial, and 0.89M temporal), which is 41.8\% fewer than TrajFlow (16.83M), the lightest competing baseline, and 82.0\% fewer than Cardiff (54.31M). Its generation times are 10.9, 8.5, and 17.3 ms per trajectory on Shenzhen, Porto, and Chengdu, respectively. Compared with TrajFlow, the fastest competing method, RoadSI achieves speedups of 2.1$\times$, 2.9$\times$, and 1.3$\times$ on the three datasets. Overall, RoadSI simultaneously achieves higher generation fidelity, faster generation, and a smaller model than diffusion-based and flow-matching-based baselines.

\subsection{Hyperparameter Analysis}
\begin{figure}[!t]
    \centering
    \Description{Pattern score sensitivity to the spatial noise factor $\lambda_{\gamma}^{\mathrm{sp}}$ and road guidance factor $\lambda_{\rho}$.}
    \includegraphics[width=0.47\linewidth]{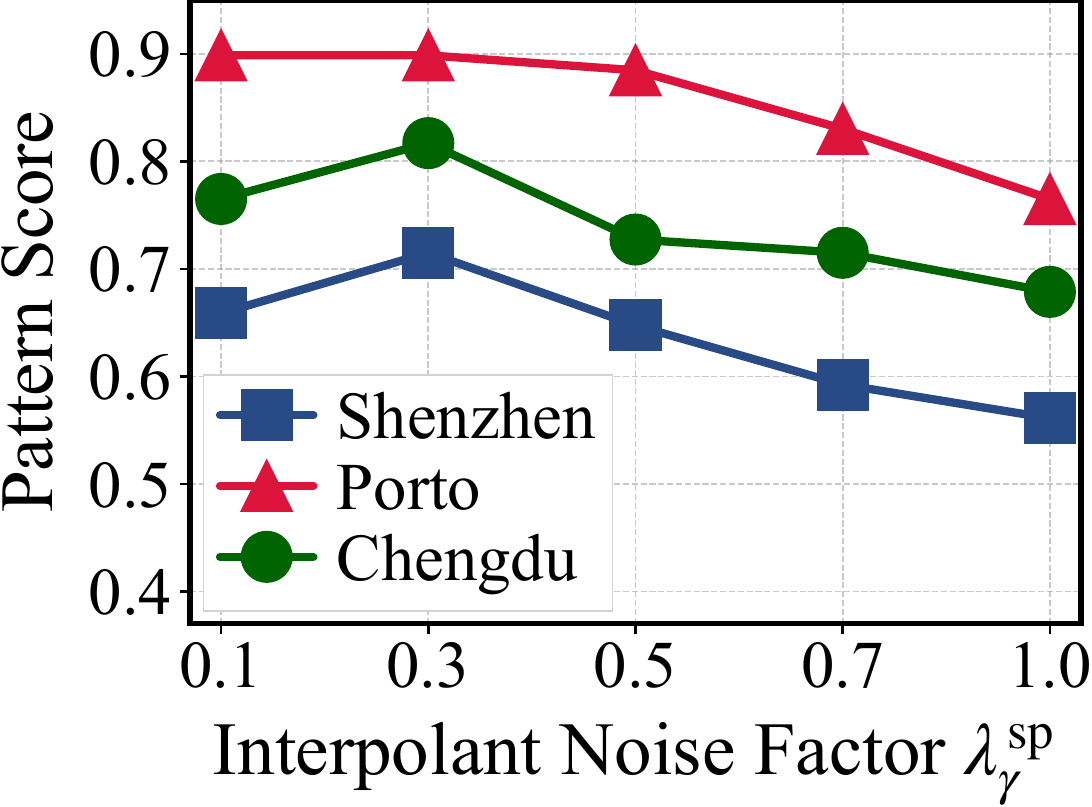}
    \quad
    \includegraphics[width=0.47\linewidth]{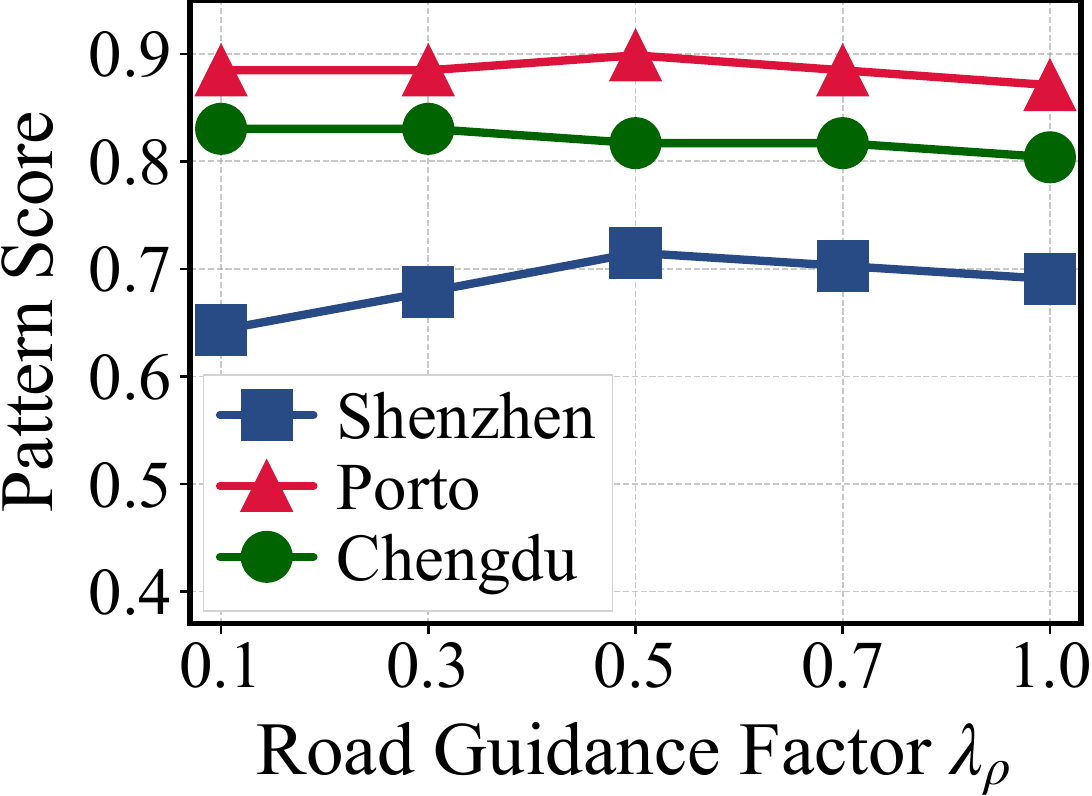}
    \caption{Pattern scores under different noise factors $\lambda_{\gamma}^{\mathrm{sp}}$ (left) and road guidance factors $\lambda_{\rho}$ (right).}
    \label{fig:lambda}
\end{figure}

We analyze two key spatial hyperparameters: the interpolant noise factor $\lambda_{\gamma}^{\mathrm{sp}}$ and the road guidance factor $\lambda_{\rho}$ by varying each over $\{0.1, 0.3, 0.5, 0.7, 1.0\}$ while fixing the other at its default value. As shown in Figure~\ref{fig:lambda}, $\lambda_{\gamma}^{\mathrm{sp}}=0.3$ achieves the highest pattern score on Shenzhen and Chengdu and matches the best result on Porto. Values above 0.3 consistently reduce pattern score, suggesting that excessive stochasticity weakens the transport signal. In contrast, RoadSI is less sensitive to $\lambda_{\rho}$: 0.5 performs best on Shenzhen and Porto and remains close to the best result on Chengdu. We therefore use $\lambda_{\gamma}^{\mathrm{sp}}=0.3$ and $\lambda_{\rho}=0.5$ by default.

\begin{figure}[!t]
  \centering
  \Description{Performance comparison of different sampling steps $K$.}
  \includegraphics[width=1\linewidth]{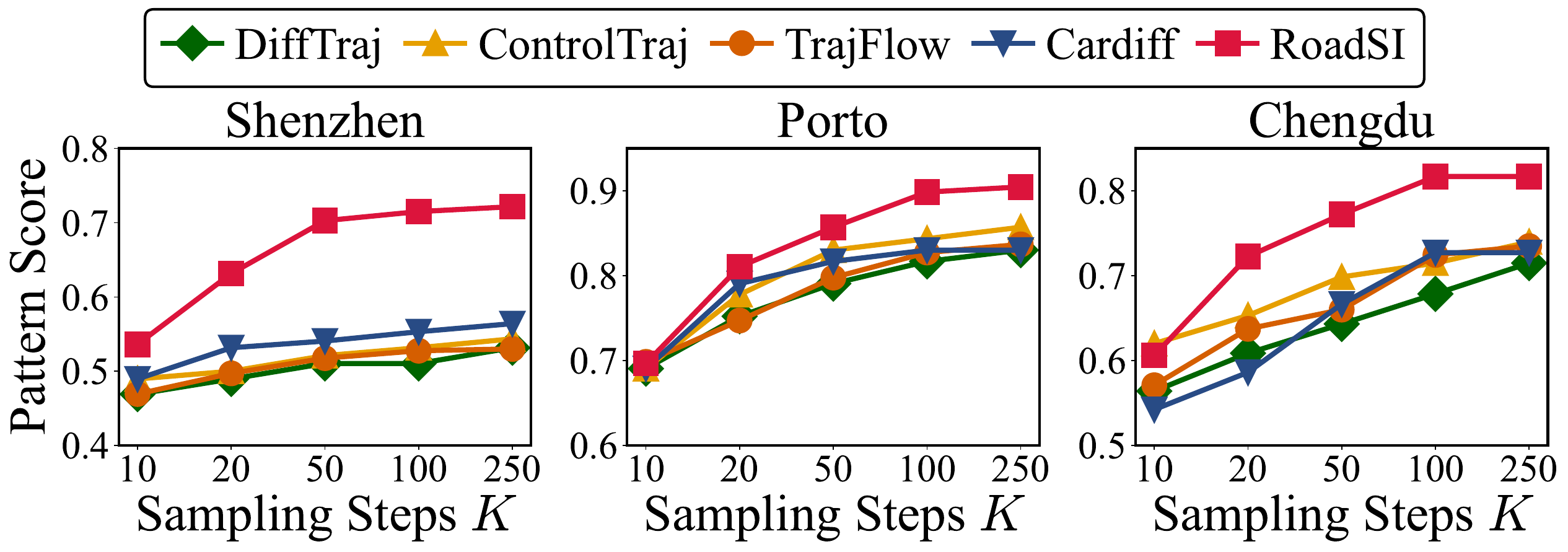}
  \caption{Pattern scores under different sampling budgets $K$.}
  \label{fig:step}
\end{figure}

We further vary the total sampling budget $K \in \{10, 20, 50, 100, 250\}$, allocating $\frac{1}{2}K$ steps to each of RoadSI's two stages, and compare against diffusion-based and flow-matching-based baselines. As shown in Figure~\ref{fig:step}, pattern score generally improves with $K$. Nevertheless, RoadSI with 20 steps already outperforms all 250-step baselines on Shenzhen, while 50 steps match or exceed their results across all three datasets, showing strong performance under a substantially smaller sampling budget.

\subsection{Downstream Utility}
To evaluate downstream utility, we consider two trajectory-analysis tasks. We partition the trajectory-generation test set into downstream training, validation, and test subsets in a $5{:}2{:}3$ ratio. We train and validate a downstream model on generated trajectories from the corresponding training and validation subsets, and evaluate it on real trajectories from the test subset. As a real-data reference, we train and validate the same model on the corresponding real subsets and evaluate it on the same real test subset.

\noindent\textbf{Destination prediction.} Given a partial trajectory, this task predicts its final destination as a continuous GPS coordinate, evaluating whether synthetic trajectories preserve long-range movement patterns. Following~\cite{rossi2018destination}, we use a GRU to encode normalized latitude-longitude prefixes and regress the destination as an offset from the last observed point. Each trajectory provides three prefixes containing its first 25\%, 50\%, and 75\% of GPS points, with the final point as the target. We report Acc@0.5km and Acc@1km, defined as the proportions of predictions within 0.5 km and 1 km of the true destination, averaged over the three observation ratios.

\begin{figure}[!t]
  \centering
  \Description{Performance on destination prediction.}
  \includegraphics[width=1\linewidth]{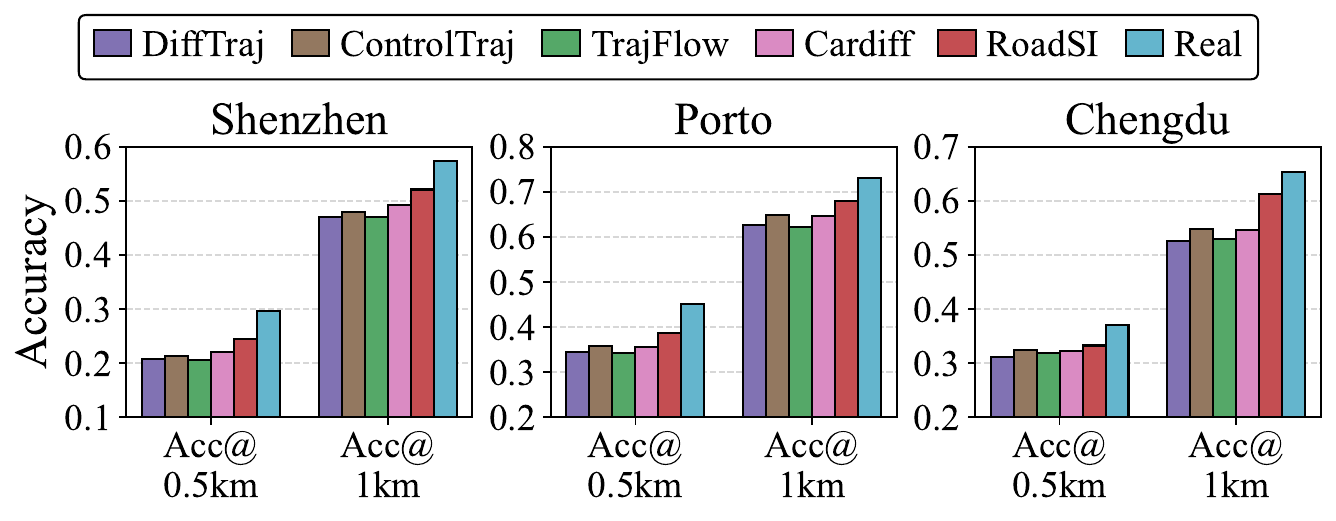}
  \caption{Performance on destination prediction.}
  \label{fig:destination}
\end{figure}

Figure~\ref{fig:destination} shows that RoadSI achieves the highest accuracy among all generative baselines across both thresholds and all datasets. Averaged over Acc@0.5km and Acc@1km, models trained on RoadSI-generated trajectories recover 88.06\%, 90.24\%, and 92.38\% of the accuracy achieved using real training data on Shenzhen, Porto, and Chengdu, respectively, while outperforming the strongest generative baseline by 7.38\%, 5.84\%, and 8.44\%. These results demonstrate that trajectories generated by RoadSI better preserve the long-range spatial cues needed for destination prediction.

\noindent\textbf{Anomalous trajectory detection.} This task evaluates whether synthetic trajectories preserve normal mobility patterns for reliable outlier detection. We partition the study area into $50\text{m} \times 50\text{m}$ grid cells and use the unsupervised GM-VSAE detector~\cite{DBLP:conf/icde/Liu0CB20}. Following~\cite{DBLP:conf/icde/Liu0CB20}, we inject detour anomalies into 5\% of the trajectories, where $\omega$ denotes the displacement in grid cells and $\nu$ the affected trajectory fraction. For example, $\omega=3; \nu=0.1$ displaces 10\% of a trajectory by three grid cells. We evaluate two anomaly settings, $\omega=3; \nu=0.1$ and $\omega=5; \nu=0.1$, and report PR-AUC.

\begin{figure}[!t]
  \centering
  \Description{Performance on anomalous trajectory detection.}
  \includegraphics[width=1\linewidth]{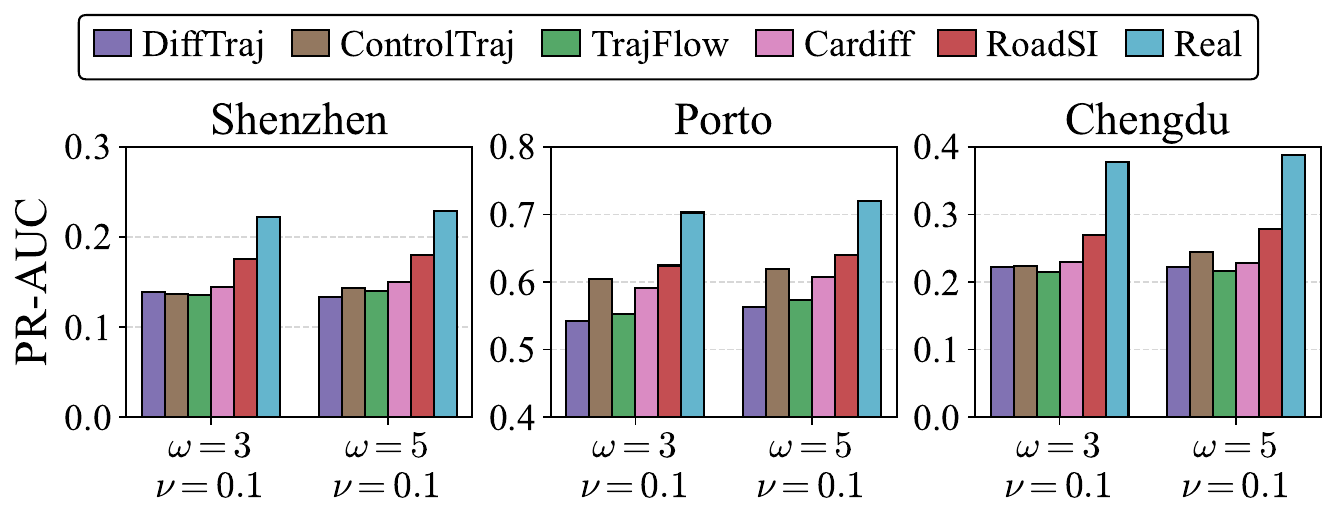}
  \caption{Performance on anomalous trajectory detection.}
  \label{fig:detection}
\end{figure}

Results in Figure~\ref{fig:detection} show that RoadSI consistently achieves the highest PR-AUC among all generative baselines across both anomaly settings and all three datasets. Averaged over the two settings, models trained on RoadSI-generated trajectories recover 78.80\%, 88.89\%, and 71.65\% of the PR-AUC achieved using real training data on Shenzhen, Porto, and Chengdu, respectively. RoadSI also improves upon the strongest competing generative baseline by 21.00\%, 3.38\%, and 16.90\% on the three datasets, respectively, showing that it better preserves mobility patterns required for anomaly detection.

\subsection{Road Adherence Evaluation}
To quantify road adherence, we report the Point-to-Road Distance Error (Point2Road) in Table~\ref{table:point2road}. It is defined as the JSD between the point-to-road distance distributions of generated and real trajectories, with lower values indicating better road adherence. RoadSI achieves the lowest error across all three datasets, demonstrating that its explicit road-guided refinement more accurately preserves the road-alignment characteristics of real trajectories.

\subsection{Effect of Re-Matching Batch Size}
Table~\ref{table:rematching} reports the runtime and matching quality under different batch configurations, where $N_s$, $N_c$, and $N_p$ denote the numbers of samples in Shenzhen, Chengdu, and Porto, respectively. Larger batches consistently improve matching quality but require more time. In Shenzhen, increasing the batch size from $\lfloor N_s/8 \rfloor$ to $N_s$ reduces the matched distance from $0.4406$ km to $0.3368$ km, while increasing the runtime. Similarly, the reduction improves in Chengdu and Porto, while the runtime rises from $106.32$ s to $703.99$ s and from $638.33$ s to $2582.93$ s, respectively. Overall, smaller batches offer substantial runtime savings at the cost of weaker coupling. Nevertheless, re-matching accounts for less than 10\% of the total training time in Shenzhen and less than 5\% in Porto and Chengdu.

\begin{table}[!t]
\centering\small
\caption{Point-to-Road Distance Error (Lower Is Better).}
\begin{tabular}{lccc}
    \toprule
    Method & Shenzhen & Porto & Chengdu \\
    \midrule
    TrajGAN     & 0.08618 & 0.08367 & 0.12570 \\
    TrajVAE     & 0.09231 & 0.23385 & 0.15774 \\
    DiffTraj    & 0.04763 & 0.04503 & 0.06216 \\
    ControlTraj & 0.07342 & 0.03012 & 0.07765 \\
    TrajFlow    & 0.04680 & 0.03013 & 0.08930 \\
    Cardiff     & 0.07525 & 0.02333 & 0.03880 \\
    RoadSI      & \textbf{0.00887} & \textbf{0.01361} & \textbf{0.01490} \\
    \bottomrule
\end{tabular}
\label{table:point2road}
\end{table}

\begin{table}[!t]
\centering\small
\caption{Effect of Re-Matching Batch Size on Efficiency.}
\resizebox{\linewidth}{!}{%
\begin{tabular}{ccccc}
    \toprule
    Dataset & Batch Size & Time (s) & Distance (km) & Reduction (\%) \\
    \midrule
    \multirow[c]{5}{*}{Shenzhen}
     & Original & -- & 0.6052 & -- \\
     & $\lfloor N_s/8 \rfloor$ & 7.98 & 0.4406 & 27.20 \\
     & $\lfloor N_s/4 \rfloor$ & 31.99 & 0.4026 & 33.47 \\
     & $\lfloor N_s/2 \rfloor$ & 162.10 & 0.3675 & 39.28 \\
     & $N_s$ & 787.17 & 0.3368 & 44.34 \\
    \midrule
    \multirow[c]{5}{*}{Porto}
     & Original & -- & 0.5216 & -- \\
     & $\lfloor N_p/28 \rfloor$ & 638.33 & 0.2679 & 48.64 \\
     & $\lfloor N_p/24 \rfloor$ & 811.25 & 0.2625 & 49.66 \\
     & $\lfloor N_p/20 \rfloor$ & 1397.36 & 0.2565 & 50.82 \\
     & $\lfloor N_p/16 \rfloor$ & 2582.93 & 0.2493 & 52.19 \\
     \midrule
    \multirow[c]{5}{*}{Chengdu}
     & Original & -- & 0.5196 & -- \\
     & $\lfloor N_c/20 \rfloor$ & 106.32 & 0.4014 & 22.74 \\
     & $\lfloor N_c/16 \rfloor$ & 154.60 & 0.3908 & 24.78 \\
     & $\lfloor N_c/12 \rfloor$ & 272.54 & 0.3774 & 27.37 \\
     & $\lfloor N_c/8 \rfloor$ & 703.99 & 0.3588 & 30.94 \\
    \bottomrule
\end{tabular}%
}
\label{table:rematching}
\end{table}

\begin{table}[!t]
\centering\small\setlength{\tabcolsep}{0.9mm}
\caption{Effect of Re-Matching Batch Size on Quality.}
\resizebox{\linewidth}{!}{%
\begin{tabular}{ccccccc}
    \toprule
    Dataset & Batch Size & Dist$\downarrow$ & StepDist$\downarrow$ & Radius$\downarrow$ & Density$\uparrow$ & Pattern$\uparrow$ \\
    \midrule
    \multirow[c]{4}{*}{Shenzhen}
     & $\lfloor N_s/8 \rfloor$ & 0.00048 & 0.00079 & 0.00023 & 0.91120 & 0.65363 \\
     & $\lfloor N_s/4 \rfloor$ & 0.00046	& 0.00074 & 0.00021	& 0.91513 & 0.67895 \\
     & $\lfloor N_s/2 \rfloor$ & 0.00041 & 0.00074 & 0.00017 & 0.92348 & 0.68502 \\
     & $N_s$ & {0.00041} & {0.00063} & {0.00016} & {0.92978} & {0.71493} \\
    \midrule
    \multirow[c]{4}{*}{Porto}
     & $\lfloor N_p/28 \rfloor$ & 0.00020 & 0.00038 & 0.00010 & 0.98593 & 0.85433 \\
     & $\lfloor N_p/24 \rfloor$ & 0.00017 & 0.00029 & 0.00009 & 0.98682 & 0.87081 \\
     & $\lfloor N_p/20 \rfloor$ & 0.00014 & 0.00029 & 0.00007 & 0.98773 & 0.87081 \\
     & $\lfloor N_p/16 \rfloor$ & {0.00015} & {0.00027} & {0.00007} & {0.99277} & {0.89855} \\
    \midrule
    \multirow[c]{4}{*}{Chengdu}
     & $\lfloor N_c/20 \rfloor$ & 0.00021 & 0.00006 & 0.00012 & 0.92685 & 0.72498 \\
     & $\lfloor N_c/16 \rfloor$ & 0.00018 & 0.00005 & 0.00009 & 0.93800 & 0.75456 \\
     & $\lfloor N_c/12 \rfloor$ & 0.00017 & 0.00003 & 0.00010 & 0.94423 & 0.75865 \\
     & $\lfloor N_c/8 \rfloor$ & {0.00016} & {0.00004} & {0.00005} & {0.95599} & {0.81690} \\
    \bottomrule
\end{tabular}%
}
\label{table:rematched_metrics}
\end{table}

Table~\ref{table:rematched_metrics} further shows that the stronger coupling obtained with larger re-matching batches generally improves generation quality. Across all three datasets, Density and Pattern increase monotonically with batch size, with Pattern rising from $0.65363$ to $0.71493$ on Shenzhen, from $0.85433$ to $0.89855$ on Porto, and from $0.72498$ to $0.81690$ on Chengdu. Dist and StepDist also generally decrease, with minor fluctuations in Porto Dist and Chengdu StepDist. These results suggest that larger batches yield better-aligned reference-real pairs and facilitate interpolant learning.

\subsection{Visualization}

\begin{figure*}[!t]
\centering
\subfloat[RoadSI generated trajectories.]{%
  \includegraphics[width=0.35\linewidth]{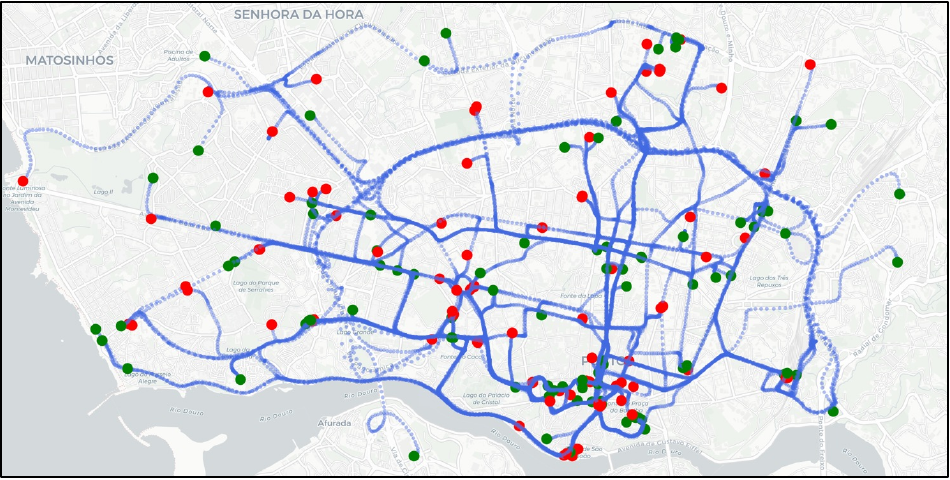}}
\qquad\quad
\subfloat[Real trajectories.]{%
  \includegraphics[width=0.35\linewidth]{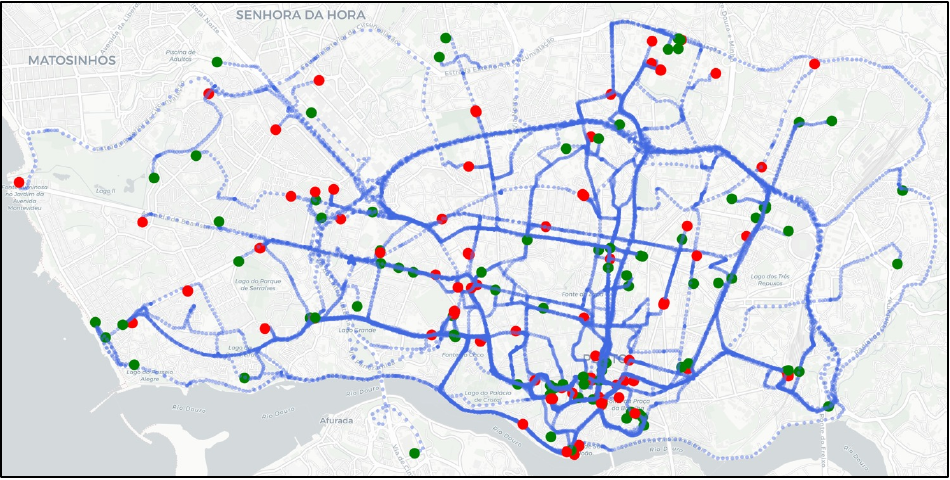}}
\caption{Visualization on the Porto dataset.}\label{fig:visual_porto}
\end{figure*}

\begin{figure*}[!t]
\centering
\subfloat[RoadSI generated trajectories.]{%
  \includegraphics[width=0.35\linewidth]{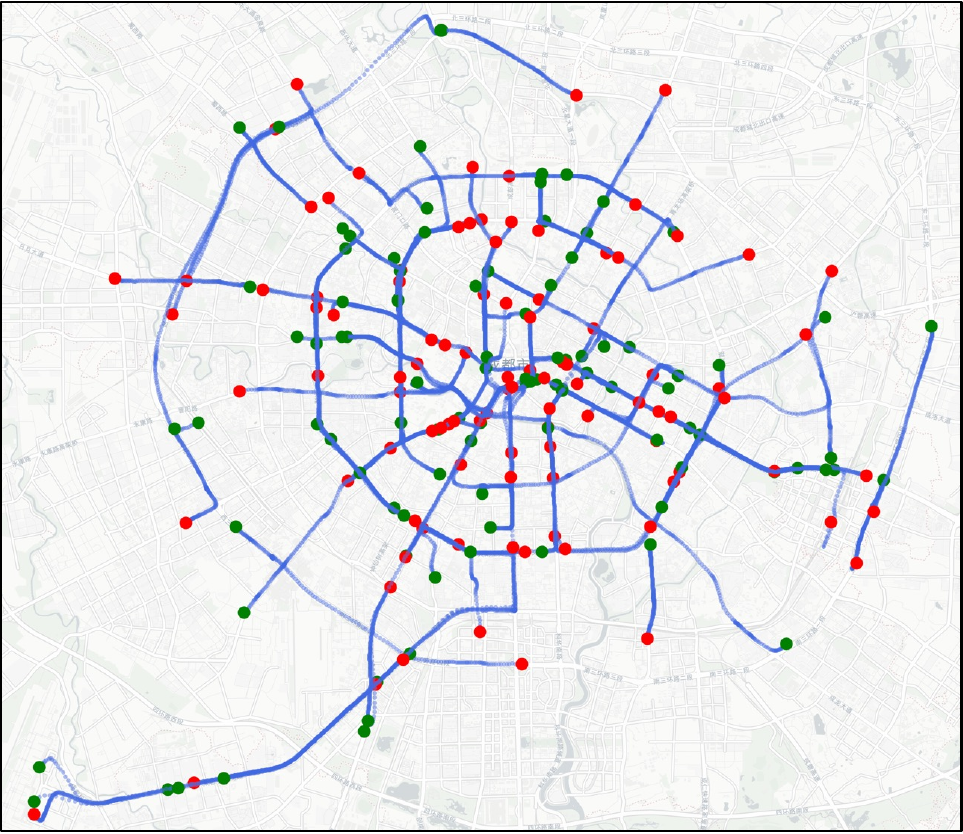}}
\qquad\quad
\subfloat[Real trajectories.]{%
  \includegraphics[width=0.35\linewidth]{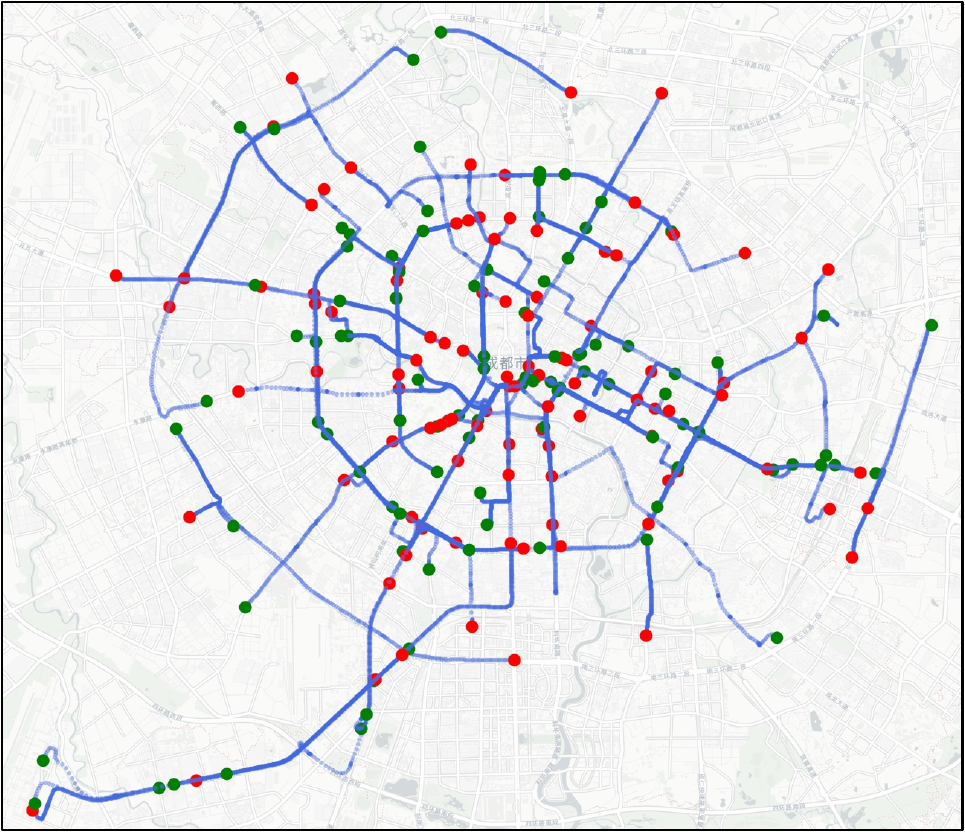}}
\caption{Visualization on the Chengdu dataset.}\label{fig:visual_cd}
\end{figure*}

\begin{figure*}[!t]
\centering
\subfloat[Reference trajectory.]{%
  \includegraphics[width=0.35\linewidth]{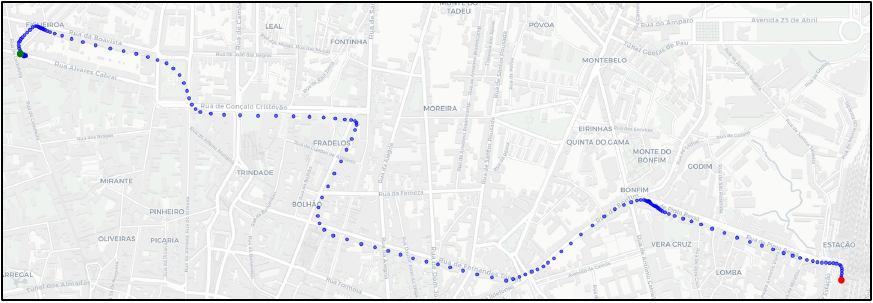}}
\qquad\quad
\subfloat[RoadSI-refined trajectory.]{%
  \includegraphics[width=0.35\linewidth]{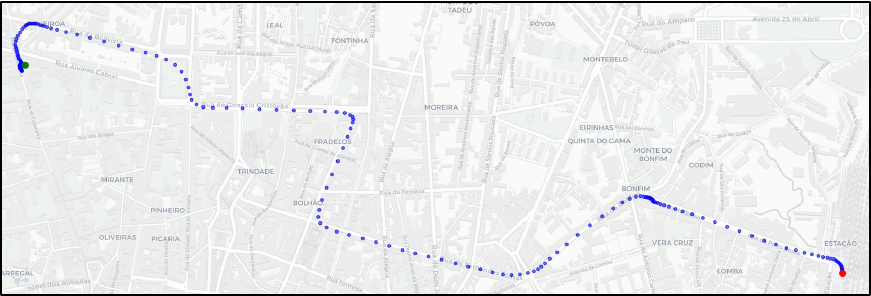}}
\caption{A case study of road-guided spatial refinement on the Porto dataset.}\label{fig:visual_porto_single}
\end{figure*}

\begin{figure*}[!t]
\centering
\subfloat[Reference trajectory.]{%
  \includegraphics[width=0.35\linewidth]{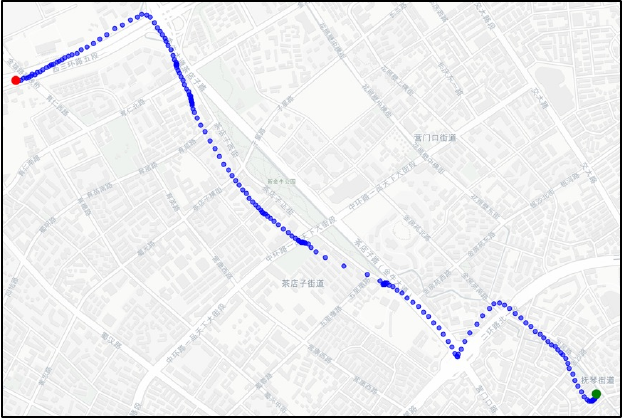}}
\qquad\quad
\subfloat[RoadSI-refined trajectory.]{%
  \includegraphics[width=0.35\linewidth]{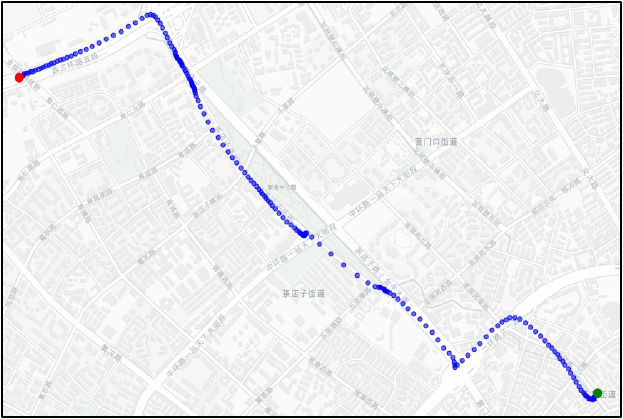}}
\caption{A case study of road-guided spatial refinement on the Chengdu dataset.}\label{fig:visual_cd_single}
\end{figure*}

To provide an intuitive assessment of generation quality, we visualize 100 trajectories generated by RoadSI and compare them with real trajectories under the same OD conditions on Porto and Chengdu (Figures~\ref{fig:visual_porto} and~\ref{fig:visual_cd}). The red and green points denote origins and destinations, respectively. Overall, RoadSI achieves geographic coverage and dominant mobility patterns similar to those of the real data, demonstrating its ability to capture spatial distribution.

Figures~\ref{fig:visual_porto_single} and~\ref{fig:visual_cd_single} further present case studies of a reference trajectory and its RoadSI-refined counterpart on Porto and Chengdu, respectively. Although the reference generator provides a useful coarse route prior, some locations deviate from the underlying road network. Road-guided refinement improves road-network conformity while preserving the overall route structure, demonstrating its effectiveness in correcting local geometric violations.

%% file: sections/6.conclusion.tex
\section{Conclusion}
We propose RoadSI, an efficient road-guided stochastic interpolant framework for spatio-temporal GPS trajectory generation. RoadSI constructs a learned spatial reference distribution and employs trajectory re-matching to establish better aligned reference-real pairs, thereby reducing transport costs and facilitating interpolant learning. It further incorporates a road score to guide spatial refinement toward road-aligned geometry and a route-conditioned temporal refinement stage to recover plausible timestamps. Experiments on three real-world datasets demonstrate that RoadSI achieves superior spatial generation quality, closely reproduces real temporal distributions, and provides the strongest downstream utility among the generative methods. With only 9.79M parameters, RoadSI also achieves faster generation than competing diffusion-based and flow-matching-based methods while maintaining its quality advantage.

%% file: sections/appendix.tex
\section{Details of Spatial Reference Trajectory Generation}
\label{ap:reference}
The reference trajectory generator $p_\theta$ is instantiated as a Denoising Diffusion Probabilistic Model (DDPM)~\cite{DBLP:conf/nips/HoJA20}. A DDPM consists of a forward diffusion process that gradually perturbs data into noise and a reverse denoising process that reconstructs data from noise. In our framework, the model is trained on real spatial trajectories $\tau_1^{\mathrm{sp}} \sim \pi_1^{\mathrm{sp}}$ conditioned on their origin--destination pairs. After training, samples generated by the model are used as spatial reference trajectories and are denoted by $\tau_0^{\mathrm{sp}}$.

\textbf{Forward process.} The forward process gradually adds Gaussian noise according to a predefined variance schedule $\{\beta_{k_i}\}_{i=1}^{K_r}$:
\begin{equation}
    q(\tau_{k_1:k_{K_r}}^{\mathrm{sp}} \mid \tau_{k_0}^{\mathrm{sp}}) = \prod_{i=1}^{K_r} q(\tau_{k_i}^{\mathrm{sp}} \mid \tau_{k_{i-1}}^{\mathrm{sp}}),
\end{equation}
\begin{equation}
    q(\tau_{k_i}^{\mathrm{sp}} \mid \tau_{k_{i-1}}^{\mathrm{sp}}) = \mathcal{N}(\tau_{k_i}^{\mathrm{sp}}; \sqrt{1-\beta_{k_i}}\tau_{k_{i-1}}^{\mathrm{sp}}, \beta_{k_i}\mathbf{I}),
\end{equation}
where $0 = k_0 < k_1 < \dots < k_{K_r}$ are discrete time steps. Here, $\tau_{k_0}^{\mathrm{sp}}$ corresponds to the clean data $\tau_1^{\mathrm{sp}}$ during training. As $k_i$ increases, the latent state $\tau_{k_i}^{\mathrm{sp}}$ approaches the standard Gaussian distribution.

\textbf{Reverse process.} The reverse process aims to recover the data from noise via a parameterized Markov chain:
\begin{equation}
    p_\theta(\tau_{k_0:k_{K_r}}^{\mathrm{sp}} \mid o, d) = p(\tau_{k_{K_r}}^{\mathrm{sp}}) \prod_{i=1}^{{K_r}} p_\theta(\tau_{k_{i-1}}^{\mathrm{sp}} \mid \tau_{k_i}^{\mathrm{sp}}, o, d),
\end{equation}
\begin{equation}
    p_\theta(\tau_{k_{i-1}}^{\mathrm{sp}} \mid \tau_{k_i}^{\mathrm{sp}}, o, d) = \mathcal{N}\big(\tau_{k_{i-1}}^{\mathrm{sp}}; \mu_\theta(\tau_{k_i}^{\mathrm{sp}}, k_i, o, d), \sigma_{k_i}^2\mathbf{I}\big),
\end{equation}
where the mean $\mu_\theta$ is re-parameterized to predict the noise $\epsilon_\theta$:
\begin{equation}
\begin{aligned}
    &\mu_\theta(\tau_{k_i}^{\mathrm{sp}}, k_i, o, d) \\
    &\quad = \frac{1}{\sqrt{\alpha_{k_i}}}\left(\tau_{k_i}^{\mathrm{sp}}-\frac{\beta_{k_i}}{\sqrt{1-\bar{\alpha}_{k_i}}}\epsilon_\theta(\tau_{k_i}^{\mathrm{sp}}, k_i, o, d)\right),
\end{aligned}
\end{equation}
where $\alpha_{k_i} = 1 - \beta_{k_i}$ and $\bar{\alpha}_{k_i} = \prod_{j=1}^{i} \alpha_{k_j}$. The variance is set to $\sigma_{k_i}^2 = \frac{1-\bar{\alpha}_{k_{i-1}}}{1-\bar{\alpha}_{k_i}}\beta_{k_i}$.

\textbf{Training objective.} The model is optimized by minimizing the Mean Squared Error (MSE):
\begin{equation}
    \mathcal{L}_{\mathrm{ref}}(\theta)
    = \mathbb{E}_{\substack{\tau_1^{\mathrm{sp}}\sim\pi_1^{\mathrm{sp}},\\ i\sim\operatorname{U}\{1,\ldots,K_r\},\\ \epsilon\sim\mathcal{N}(\mathbf{0},\mathbf{I})}}
    \left[ \left\| \epsilon - \epsilon_\theta(\tau_{k_i}^{\mathrm{sp}}, k_i, o, d) \right\|^2 \right].
\end{equation}
where $\tau_{k_i}^{\mathrm{sp}} = \sqrt{\bar{\alpha}_{k_i}}\tau_1^{\mathrm{sp}} + \sqrt{1-\bar{\alpha}_{k_i}}\epsilon$.

\textbf{Reference trajectory generation.} After training, we generate the reference spatial trajectory $\tau_0^{\mathrm{sp}}$ by starting from noise $\tau_{k_{K_r}}^{\mathrm{sp}} \sim \mathcal{N}(\mathbf{0}, \mathbf{I})$ and performing reverse sampling. We employ DDIM~\cite{DBLP:conf/iclr/SongME21} during sampling to improve the efficiency.

\section{Score Component Derivations}
In this part, we present the detailed derivations for the data score $s_{\mathrm{data}}$ and the road score $s_{\mathrm{road}}$.
\subsection{Data Score Derivation}
\label{ap:data_score}
We derive the relationship between the data score and the noise component of the spatial stochastic interpolant. Recall that
\begin{equation}
\tilde{\tau}_t^{\mathrm{sp}}=(1-t)\tilde{\tau}_0^{\mathrm{sp}} + t\tilde{\tau}_1^{\mathrm{sp}} + \gamma_{\mathrm{sp}}(t)z,
\end{equation}
where $(\tilde{\tau}_0^{\mathrm{sp}}, \tilde{\tau}_1^{\mathrm{sp}}) \sim \pi_{0,1}^{\mathrm{sp}}$, and $z\sim\mathcal{N}(\mathbf{0},\mathbf{I})$.

Let $I_t = (1-t)\tilde{\tau}_0^{\mathrm{sp}} + t\tilde{\tau}_1^{\mathrm{sp}}$ denote the linear interpolation term. Conditioned on the endpoint pair, the interpolant follows
\begin{equation}
p_t \left(x \mid \tilde{\tau}_0^{\mathrm{sp}}, \tilde{\tau}_1^{\mathrm{sp}} \right) =
\mathcal{N} \left(x;I_t, \gamma_{\mathrm{sp}}^2(t)\mathbf{I} \right),
\end{equation}
where $x$ denotes an arbitrary intermediate trajectory state.

The marginal density at time $t$ is obtained by integrating over the endpoint coupling:
\begin{equation}
p_t(x)=\int p_t\left(x\mid\tilde{\tau}_0^{\mathrm{sp}},\tilde{\tau}_1^{\mathrm{sp}}\right)
\pi_{0,1}^{\mathrm{sp}}\left(\tilde{\tau}_0^{\mathrm{sp}},\tilde{\tau}_1^{\mathrm{sp}}\right)\mathrm{d}\tilde{\tau}_0^{\mathrm{sp}}\mathrm{d}\tilde{\tau}_1^{\mathrm{sp}}.
\end{equation}

The data score is defined as
\begin{equation} 
s_{\mathrm{data}}(t,x)=\nabla_x\log p_t(x).
\end{equation}

Expanding the gradient gives
\begin{equation}
\begin{aligned}
s_{\mathrm{data}}(t,x)&=\frac{1}{p_t(x)}\nabla_x p_t(x)\\
&=\frac{1}{p_t(x)}\int\nabla_xp_t\left(x\mid\tilde{\tau}_0^{\mathrm{sp}},\tilde{\tau}_1^{\mathrm{sp}}\right) \\
&\qquad \pi_{0,1}^{\mathrm{sp}}
\left(\tilde{\tau}_0^{\mathrm{sp}},\tilde{\tau}_1^{\mathrm{sp}}\right)\mathrm{d}\tilde{\tau}_0^{\mathrm{sp}}\mathrm{d}\tilde{\tau}_1^{\mathrm{sp}}.
\end{aligned}
\end{equation}

Since the conditional density is Gaussian,
\begin{equation}
\begin{aligned}
    \nabla_xp_t\left(x\mid\tilde{\tau}_0^{\mathrm{sp}},\tilde{\tau}_1^{\mathrm{sp}}\right)
    &=-\frac{x-I_t}{\gamma_{\mathrm{sp}}^2(t)}
p_t\left(x\mid\tilde{\tau}_0^{\mathrm{sp}},\tilde{\tau}_1^{\mathrm{sp}}\right) \\
    &=-\frac{z}{\gamma_{\mathrm{sp}}(t)}
p_t\left(x\mid\tilde{\tau}_0^{\mathrm{sp}},\tilde{\tau}_1^{\mathrm{sp}}\right).
\end{aligned}
\end{equation}

Plugging this back into the integral yields:
\begin{equation}
\begin{aligned}
s_{\mathrm{data}}(t,x)
&= \frac{1}{p_t(x)} \int - \frac{z}{\gamma_{\mathrm{sp}}(t)} p_t(x \mid \tilde{\tau}_0^{\mathrm{sp}}, \tilde{\tau}_1^{\mathrm{sp}}) \\
&\qquad \pi_{0,1}^{\mathrm{sp}}(\tilde{\tau}_0^{\mathrm{sp}}, \tilde{\tau}_1^{\mathrm{sp}}) \mathrm{d}\tilde{\tau}_0^{\mathrm{sp}} \mathrm{d}\tilde{\tau}_1^{\mathrm{sp}} \\
&= - \gamma_{\mathrm{sp}}^{-1}(t) \int z \frac{p_t(x \mid \tilde{\tau}_0^{\mathrm{sp}}, \tilde{\tau}_1^{\mathrm{sp}}) \pi_{0,1}^{\mathrm{sp}}(\tilde{\tau}_0^{\mathrm{sp}}, \tilde{\tau}_1^{\mathrm{sp}})}{p_t(x)} \\
&\qquad \mathrm{d}\tilde{\tau}_0^{\mathrm{sp}} \mathrm{d}\tilde{\tau}_1^{\mathrm{sp}} \\
&= - \gamma_{\mathrm{sp}}^{-1}(t) \mathbb{E}[z \mid \tilde{\tau}_t^{\mathrm{sp}}=x].
\end{aligned}
\end{equation}

The optimal noise component is defined as
$\eta(t,x)=\mathbb{E}[z\mid\tilde{\tau}_t^{\mathrm{sp}}=x]$.
Accordingly, the data score can be expressed as
\begin{equation}
    s_{\mathrm{data}}(t,x)
    =
    -\gamma_{\mathrm{sp}}^{-1}(t)\eta(t,x).
\end{equation}

\begin{figure*}[!t]
    \centering
    \Description{Architecture of reference trajectory generation network.}
    \includegraphics[width=0.75\linewidth]{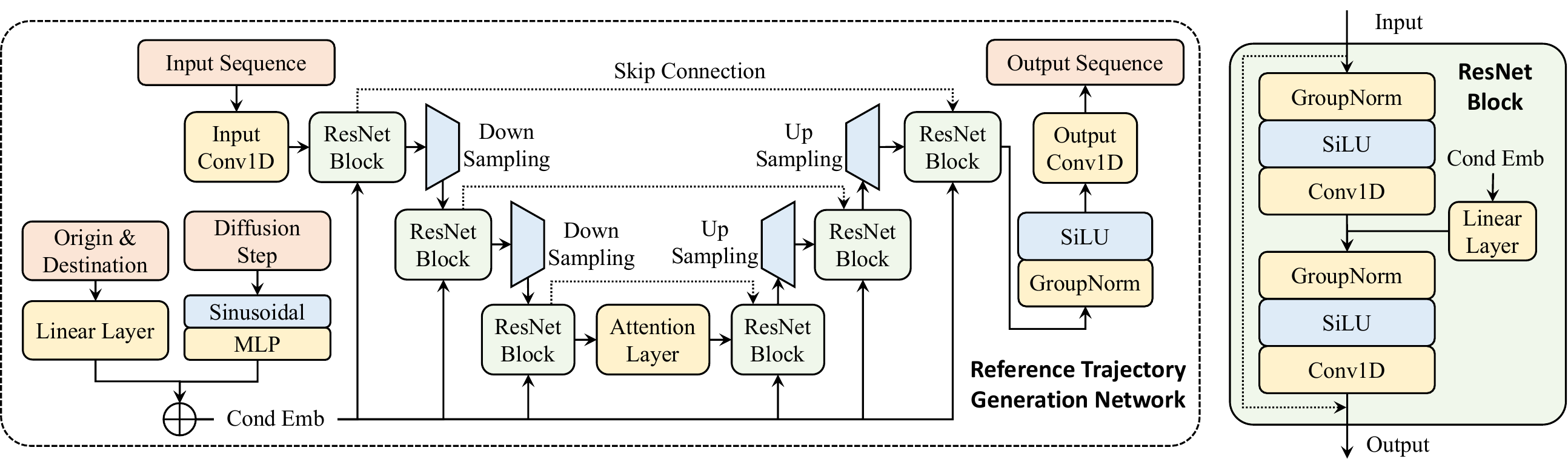}
    \caption{Architecture of the reference trajectory generation network.}
    \label{fig:reference_net}
\end{figure*}

\begin{figure*}[!t]
    \centering
    \Description{Architecture of drift prediction network and noise prediction network.}
    \includegraphics[width=1\linewidth]{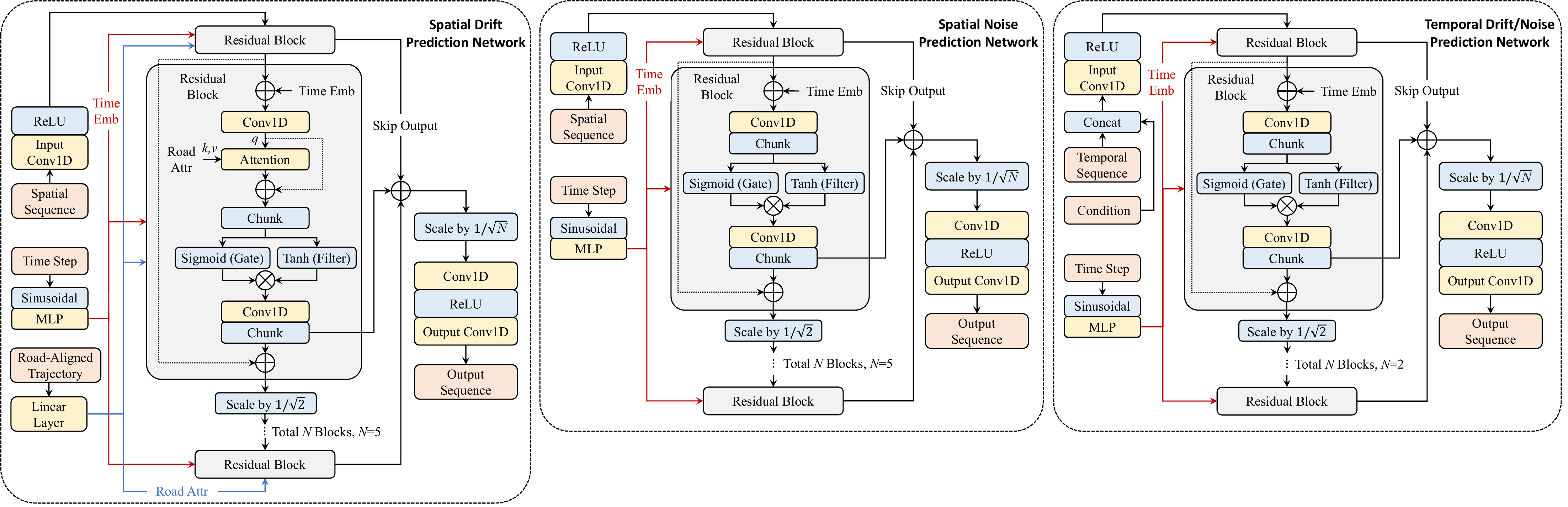}
    \caption{Architectures of the drift and noise prediction networks.}
    \label{fig:dn_nets}
\end{figure*}

\subsection{Road Score Derivation}
\label{ap:road_score}
We provide the detailed derivation of the road score $s_{\mathrm{road}}$ for a specific spatial point $\tilde{\ell}_t^{(i)}$. Let $d_t^{(i)} = \|\tilde{\ell}_t^{(i)} - r(\tilde{\ell}_t^{(i)})\|$ denote the Euclidean distance to the road projection. The point-wise potential function is defined as:
\begin{equation}
    U_{\mathrm{road}}\bigl(t, \tilde{\ell}_t^{(i)}, r(\tilde{\ell}_t^{(i)})\bigr) = \mathrm{softplus}\bigl(d_t^{(i)} - R(t)\bigr).
\end{equation}

Treating $r(\cdot)$ as a non-differentiable projection, the road score is defined as the negative gradient of the potential with respect to $\tilde{\ell}_t^{(i)}$. Applying the chain rule, we obtain:
\begin{equation}
    s_{\mathrm{road}}\bigl(t, \tilde{\ell}_t^{(i)}, r(\tilde{\ell}_t^{(i)})\bigr) = -\nabla_{\tilde{\ell}_t^{(i)}} U_{\mathrm{road}}= -\frac{\partial U_{\mathrm{road}}}{\partial d_t^{(i)}} \cdot \nabla_{\tilde{\ell}_t^{(i)}} d_t^{(i)}.
\end{equation}

The first term corresponds to the derivative of the softplus function, which is the sigmoid function $\sigma(\cdot)$:
\begin{equation}
\begin{aligned}
    \frac{\partial U_{\mathrm{road}}}{\partial d_t^{(i)}}
    &= \frac{\partial}{\partial d_t^{(i)}} \log\bigl(1 + e^{d_t^{(i)} - R(t)}\bigr) \\
    &= \frac{e^{d_t^{(i)} - R(t)}}{1 + e^{d_t^{(i)} - R(t)}} = \sigma\bigl(d_t^{(i)} - R(t)\bigr).
\end{aligned}
\end{equation}

The second term constitutes the gradient of the Euclidean distance:
\begin{equation}
   \nabla_{\tilde{\ell}_t^{(i)}} d_t^{(i)} = \frac{\tilde{\ell}_t^{(i)} - r(\tilde{\ell}_t^{(i)})}{d_t^{(i)}}.
\end{equation}

Substituting these terms back into the chain rule expression, we obtain the final formulation:
\begin{equation}
    s_{\mathrm{road}}\bigl(t, \tilde{\ell}_t^{(i)}, r(\tilde{\ell}_t^{(i)})\bigr) = - \sigma\bigl(d_t^{(i)} - R(t)\bigr) \cdot \frac{\tilde{\ell}_t^{(i)} - r(\tilde{\ell}_t^{(i)})}{d_t^{(i)}}.
\end{equation}

\section{Network Architectures}
\label{ap:architectures}

\textbf{Reference trajectory generation network.} For $p_\theta$, we adopt a lightweight 1D U-Net adapted from the Traj-UNet in DiffTraj~\cite{DBLP:conf/nips/ZhuYZZY23}, as shown in Figure~\ref{fig:reference_net}. The network follows an encoder–decoder design with ResNet blocks and optional down/up sampling. A self-attention layer is inserted at the bottleneck, and skip connections concatenate encoder and decoder features at corresponding resolutions. We condition the model on the diffusion step and origin–destination information. The diffusion step is encoded using sinusoidal embeddings followed by an MLP, while the concatenated origin–destination vector is mapped by a linear layer to the same dimensionality. The resulting embeddings are summed (denoted as Cond Emb in Figure~\ref{fig:reference_net}) and injected into every ResNet block. Each ResNet block contains two (GroupNorm $\rightarrow$ SiLU $\rightarrow$ Conv1D) layers, with the conditional embedding linearly projected and added after the first convolution. Down sampling is implemented via a strided 1D convolution, while up sampling uses nearest-neighbor interpolation followed by a 1D convolution. Finally, a GroupNorm–SiLU–Conv1D head maps features back to the trajectory space.

\noindent\textbf{Spatial drift prediction network.} For $v_\phi$, we design a WaveNet-style 1D residual architecture characterized by gated activation units and cross-attention mechanisms. As illustrated in Figure~\ref{fig:dn_nets} (Left), the interpolant time step is transformed into a temporal embedding via sinusoidal positional encodings and an MLP similar to $p_\theta$. To incorporate road network context, a road-aligned trajectory is processed through a linear layer and integrated into the residual blocks. Each residual block further employs an attention layer that fuses road attributes with the intermediate features. The core transformation consists of Conv1D layers followed by a gated activation (Sigmoid gate and Tanh filter). The skip outputs from all $N$ blocks are aggregated and normalized by $1/\sqrt{N}$, then passed through a ReLU-activated convolutional head to produce the final output sequence.

\noindent\textbf{Spatial noise prediction network.} The network $\eta_\delta$ is a simplified variant of $v_\phi$ sharing the fundamental WaveNet-like backbone, including the sinusoidal interpolant-time embedding and gated residual blocks. It omits the road-aligned trajectory input and the cross-attention layers. Instead, it processes the input sequence through a series of $N$ residual blocks, where each stage refines the feature map using Conv1D layers and gated activations. Similar to the drift network, the skip outputs are summed and normalized by $1/\sqrt{N}$ before the final convolutional head produces the predicted noise sequence. This architectural simplification allows $\eta_\delta$ to focus on capturing local stochastic variations in the trajectory data.

\noindent\textbf{Temporal drift/noise prediction network.} The temporal drift network $v_\xi$ and temporal noise network $\eta_\zeta$ use the same WaveNet-style residual backbone as the spatial prediction networks, with separate parameters for the two fields. They operate on the one-dimensional log-pace state $y_t$ and concatenate the route condition $h$ at the input layer. Unlike the spatial drift network, they do not use road-aligned trajectory inputs or cross-attention modules.

\section{Road-projection Complexity}
During each spatial interpolant step, road projection is computed point-wise. For each trajectory point, we retrieve a candidate road segment and project the point onto its closest location. A brute-force search over all $|E|$ road segments requires $O(|E|)$ operations per point and $O(L|E|)$ per trajectory per step, since projection onto a selected segment is $O(1)$. In practice, we use a KD-tree-based spatial index to accelerate candidate-segment retrieval, reducing the average cost to $O(L\log |E|)$ per trajectory per step. The resulting road-projection overhead is therefore $O(L\log |E|)$ per trajectory per step.